\documentclass[journal]{IEEEtran}
\usepackage{amsmath,amsfonts}
\usepackage{algorithmic}
\usepackage{algorithm}
\usepackage{array}
\usepackage[caption=false,font=normalsize,labelfont=sf,textfont=sf]{subfig}
\usepackage{textcomp}
\usepackage{stfloats}
\usepackage{url}
\usepackage{verbatim}
\usepackage{graphicx}
\usepackage{CJKutf8}
\usepackage{balance}
\usepackage{cite}

\usepackage{booktabs}
\usepackage{multirow}
\usepackage{makecell}
\usepackage{threeparttable}
\usepackage{tabularx}
\usepackage{placeins}

\usepackage{hyperref}
\usepackage{xcolor}


\begin{document}

\title{TeX-1500: A Paired Real-World LWIR Hyperspectral Dataset and Benchmark for Temperature--Emissivity--Texture Decomposition}

\author{
    Cheng Dai, Jiale Lin, Hongyi Xu, Bingxuan Song, Ziyang Xie, and Fanglin Bao
    \thanks{\emph{TeX-1500} dataset is at \url{https://huggingface.co/datasets/jialelin2007/TeX-1500}, and the TeX-1500 benchmark protocol, TeX-UNet code and pretrained models are at \url{https://github.com/dccc2025/TeX-1500}.}
    \thanks{Cheng Dai, Hongyi Xu, Ziyang Xie and Fanglin Bao are with the School of Science, Westlake University, Hangzhou 310030, China.}
    \thanks{Jiale Lin and Bingxuan Song are with the School of Engineering, Westlake University, Hangzhou 310030, China.}
    \thanks{E-mails: \{daicheng, linjiale, xuhongyi, songbingxuan, xiezhiyang, baofanglin\}@westlake.edu.cn.}
    \thanks{Cheng Dai and Jiale Lin contributed equally to this work.}
    \thanks{Corresponding author: Fanglin Bao.}
}

\markboth{preprint ~2026}
{C. Dai \MakeLowercase{\textit{et al.}}: TeX-1500: A Paired LWIR HSI--TeX Dataset and Benchmark}

\maketitle

\begin{abstract}
    Temperature--emissivity--texture (TeX) decomposition seeks to recover object heat state, material spectral response, and visible-like geometric texture from long-wave infrared hyperspectral imaging (LWIR HSI). Existing TeX pipelines are mainly scene-specific inverse solvers, and the lack of paired LWIR HSI--TeX supervision has limited learning-based decomposition. To address this gap, we introduce \emph{TeX-1500}, a large-scale paired LWIR HSI--TeX dataset and benchmark for supervised HSI-to-TeX decomposition. \emph{TeX-1500} contains 1,522 calibrated real-scene pairs from DARPA Invisible Headlights (DARPA IH) pushbroom imagery and our FTIR acquisitions, covering five locations, four seasons, diverse acquisition times, heterogeneous wavelength layouts, and two sensor families. Each sample stores a calibrated valid-band radiance cube, calibrated wavelength positions, and aligned temperature, emissivity, and texture supervision constructed through a consistent restoration and TeX-construction protocol. We further provide TeX-UNet, a simple wavelength-aware baseline that maps calibrated HSI bands and wavelength positions to TeX fields. Experiments on the held-out DARPA IH pushbroom scenes and zero-/few-shot transfer to FTIR scenes show that \emph{TeX-1500} provides usable paired supervision and a measurable benchmark for data-driven physical-property-centered thermal perception.
\end{abstract}


\section{Introduction}

\IEEEPARstart{M}{odern} visual perception relies heavily on RGB, depth, and infrared sensing, but these measurements often describe scene appearance rather than physical quantities. RGB depends on reflected illumination and is sensitive to lighting and atmospheric changes~\cite{wasserstein_all_in_one_image_restoration_2026_tpami, he_dehazing_2010_tpami, vivnet_all_in_one_image_restoration_2026_tpami}; depth sensors can fail on missing geometry, reflective surfaces, transparent objects, and adverse weather~\cite{BaoHeatassistedDetectionRanging2023, BijelicSeeingThroughFog2020,ParkRethinkingLiDARWeather2024, nature_machine_intelligence_2022_autonomous_driving,2021_prl_xufeihu_non_line_of_sight_imaging,ye2024real_yuanxin_xufeihu_ncs_2024}; and infrared images mix object self-emission, material response, and environmental radiation~\cite{BaoWhyThermalImages2024,dai2026hadarbasedthermalinfraredhyperspectral,xu2026universalcomputationalthermalimaging}. \emph{These limitations arise because conventional visual signals are sensing proxies: they describe how a scene appears to a sensor, rather than directly representing the object's stable physical properties.}

TeX decomposition aims to replace this appearance-centered view with a physical-property-centered representation. TeX separates temperature $T$, emissivity $e$, and geometric texture $X$, which respectively describe heat state, material spectral response, and scene-dependent lighting structure. HADAR-SGD~\cite{BaoHeatassistedDetectionRanging2023} and HADAR-SLOT~\cite{xu2026universalcomputationalthermalimaging} have shown the promise of recovering TeX from thermal measurements. However, existing TeX pipelines are largely scene-specific inverse solvers whose performance depends on grouping, initialization, regularization, and iterative optimization settings. This makes them valuable for constructing individual TeX reconstructions, but insufficient as a scalable training and evaluation protocol for learning-based HSI-to-TeX decomposition.

A key missing component for learning-based TeX decomposition is paired supervision across real scenes and sensors. Such supervision requires calibrated LWIR hyperspectral inputs aligned with temperature, emissivity, and texture labels, rather than isolated TeX reconstructions from individual scenes. Constructing these pairs is nontrivial for real LWIR HSI: raw measurements may contain corrupted bands, stripe artifacts, stochastic noise, wavelength shifts, sensor-dependent valid-band layouts, and scene-dependent sky radiation. Without a consistent restoration protocol, these acquisition artifacts can be absorbed into the TeX labels and become spurious supervision for neural models.

This paper presents \emph{TeX-1500}, a paired LWIR HSI--TeX dataset and initial benchmark designed to support supervised HSI-to-TeX learning and evaluation. We construct \emph{TeX-1500} from DARPA IH pushbroom imagery~\cite{YellinConcurrentBandSelection2024} and our FTIR acquisitions using a consistent construction protocol, producing calibrated HSI--TeX pairs across locations, seasons, acquisition times, wavelength layouts, and sensor families. The resulting dataset enables both within-sensor evaluation on held-out DARPA IH secnes and cross-sensor evaluation through zero-/few-shot transfer to FTIR scenes.

Our contributions are summarized as follows:
\begin{itemize}
    \item We release \emph{TeX-1500}, a paired LWIR HSI--TeX dataset containing 1,522 real-scene samples from DARPA IH pushbroom data and our FTIR acquisitions.
    \item We provide a consistent construction protocol that converts restored and wavelength-calibrated LWIR HSI into aligned temperature, emissivity, and texture supervision.
    \item We establish an initial learning-based benchmark with TeX-UNet, using its prediction performance on held-out DARPA IH test split scenes and zero-/few-shot transfer to FTIR scenes to demonstrate the learnability of \emph{TeX-1500}.
\end{itemize}

\section{Related Work}

\subsection{Datasets for Thermal Perception}

Existing spectral and thermal datasets have enabled many learning-based tasks by exposing information beyond broadband RGB, but most public benchmarks are not designed for thermal physical-property estimation.

VIS--NIR hyperspectral datasets such as CAVE~\cite{cave_dataset_2010_tip}, Harvard~\cite{havard_2011_cvpr}, ICVL~\cite{icvl_2016_eccv}, ARAD-1K~\cite{arad_ntire_2022_cvpr}, Houston~\cite{huston_2013_jstars}, and WHU-Hi~\cite{whu_hi_2020_arxiv} have supported spectral reconstruction, remote-sensing classification, multimodal fusion, and low-light perception~\cite{satmae_2022_nips, spectralgpt_2024_tpami, spectralearth_2025_jstars, hyperSIGMA_2025_tpami}. Their measurements, however, are dominated by reflected solar or artificial illumination. As a result, they provide rich appearance spectra but do not carry the long-wave self-emission and material-response cues needed for temperature--emissivity--texture decomposition~\cite{BaoHeatassistedDetectionRanging2023}.

Visible--infrared paired datasets move closer to thermal perception, but they provide broadband thermal images rather than wavelength-resolved LWIR radiance. KAIST~\cite{KAIST_2015_cvpr}, CVC14~\cite{cvc14_sensors_2016}, TNO~\cite{tno_2017}, MFNet~\cite{mfnet_2017_iros}, Freiburg~\cite{freiburg_2020_iros}, LLVIP~\cite{LLVIP_2021_cvpr}, Boson-Night~\cite{boson_nighttime_2023_iros}, and MSRS~\cite{piafusion_2022_inf} have supported detection, segmentation, image fusion, and low-light enhancement by combining visible appearance with thermal contrast~\cite{Mask_Difuser_2026_tpami, ddfm_2023_iccv, thermalgen_2026_nips}. These benchmarks are valuable for thermal appearance modeling; however, their broadband thermal channel collapses the spectral variation needed to disentangle temperature, emissivity, and texture~\cite{BaoHeatassistedDetectionRanging2023, xu2026universalcomputationalthermalimaging, dai2026hadarbasedthermalinfraredhyperspectral}.

LWIR hyperspectral datasets provide the spectral measurements needed for this physical decomposition, yet paired TeX supervision remains missing. Existing Telops datasets~\cite{ieee_grss_datasets_2015_jstars, wangdu2025spectral_igarss, wangdu_isprs_2026_tes} and DARPA IH dataset~\cite{YellinConcurrentBandSelection2024} contain real long-wave spectral imagery and have supported tasks such as target analysis and thermal remote sensing. Their public releases, however, do not provide calibrated ground-based LWIR HSI paired with aligned temperature, emissivity, and texture fields. \emph{TeX-1500} addresses this dataset gap by releasing calibrated HSI--TeX supervisions across locations, seasons, acquisition times, wavelength layouts, and sensor families.

\subsection{LWIR HSI Applications and TeX Solvers}

Beyond datasets, prior LWIR HSI methods show that long-wave spectra contain physical information useful for inference. LWIR HSI has been used for passive ranging, classification, target analysis, thermal retrieval, and physically guided reconstruction~\cite{DorkenGallastegiAbsorptionBasedPassiveRange2025, ieee_grss_datasets_2015_jstars, CaoLWIRHyperspectralImageClassification2022, YellinConcurrentBandSelection2024, wangdu2025spectral_igarss, wangdu_isprs_2026_tes, MousaPhysicsIntegratedInference2026}. Recent infrared generation, neural rendering, and compressive HSI methods also introduce temperature- or material-aware priors~\cite{mao2026pid,zhao2026thermal,meng2025pcmamba}. Together, these studies show that LWIR spectra encode physical cues beyond broadband thermal contrast. However, they typically use these cues for a downstream task or as intermediate priors, rather than providing complete temperature, emissivity, and texture targets paired with LWIR HSI measurements.

Model-based TeX solvers make this physical decomposition explicit. HADAR-SGD~\cite{BaoHeatassistedDetectionRanging2023} estimates temperature, emissivity, and texture by fitting a thermal rendering equation with a material library; HADAR-SLOT~\cite{xu2026universalcomputationalthermalimaging} removes the library assumption through autonomous estimation of emissivity~\cite{TsengBlockCoordinateDescent2001}; and HAIR~\cite{dai2026hadarbasedthermalinfraredhyperspectral} couples the HADAR rendering equation with downwelling radiative transfer to restore degraded LWIR HSI. These solvers establish the physical basis of TeX decomposition, but their per-scene optimization depends on choices such as grouping, initialization, regularization, and optimization settings. \emph{TeX-1500} complements this line of work by converting TeX decomposition from per-scene optimization into a paired learning benchmark with calibrated LWIR HSI inputs, constructed TeX supervision, and fixed evaluation settings for neural baselines such as TeX-UNet.

\section{Dataset}

\begin{table*}[t]
    \centering
    \caption{Dataset composition of TeX-1500.}
    \label{tab:dataset_summary}
    \scriptsize
    \setlength{\tabcolsep}{4pt}
    \renewcommand{\arraystretch}{1.12}
    \resizebox{\textwidth}{!}{%
        \begin{tabular}{@{}l>{\raggedright\arraybackslash}p{3.55cm}cc>{\raggedright\arraybackslash}p{1.85cm}p{2.05cm}p{2.45cm}c@{}}
            \toprule
            Split & Location                                    & Scenes & Images & Dates            & Wavelengths ($\mu$m)    & Spatial size                  & Bands                    \\
            \midrule
            \multicolumn{8}{@{}l}{\textit{DARPA IH pushbroom subset}}                                                                                                                     \\
            \addlinespace[1pt]
            Train & \makecell[l]{Sidewinder Range, TPG, AZ}     & 12     & 204    & 2021.08          & \makecell[c]{8.1--13.2} & \makecell[c]{$260\times1500$} & \makecell[c]{256}        \\
            Train & \makecell[l]{Loring Commerce Center, ME}    & 44     & 404    & 2021.12          & \makecell[c]{6.8--13.1                                                             \\8.0--13.1} & \makecell[c]{$480\times1700$\\$260\times1800$} & \makecell[c]{250\\256} \\
            Train & \makecell[l]{Avon Park Air Force Range, FL} & 18     & 488    & 2022.04          & \makecell[c]{6.8--13.1                                                             \\8.0--13.1} & \makecell[c]{$260\times1200$\\$480\times1700$} & \makecell[c]{250\\256} \\
            Valid & \makecell[l]{Sidewinder Range, TPG, AZ}     & 8      & 51     & 2020.09          & \makecell[c]{8.1--13.2} & \makecell[c]{$260\times1280$} & \makecell[c]{256}        \\
            Test  & \makecell[l]{Fort A. P. Hill, VA}           & 18     & 233    & 2021.04          & \makecell[c]{8.1--13.2} & \makecell[c]{$260\times1600$} & \makecell[c]{256}        \\
            \midrule
            \multicolumn{8}{@{}l}{\textit{FTIR subset}}                                                                                                                                   \\
            \addlinespace[1pt]
            Train & \makecell[l]{Wuhan University, China}       & 38     & 111    & 2024.02--2025.11 & \makecell[c]{7.9--11.5} & \makecell[c]{$320\times256$}  & \makecell[c]{86}         \\
            Test  & \makecell[l]{Wuhan University, China}       & 16     & 31     & 2026.01          & \makecell[c]{7.9--11.5} & \makecell[c]{$320\times256$}  & \makecell[c]{86/124/277} \\
            \bottomrule
        \end{tabular}%
    }

    \vspace{2pt}
    \begin{minipage}{\textwidth}
        \raggedright
        \scriptsize
        \textit{Note.} The wavelength ranges are nominal wavelengths. The
        statistics summarize the main distribution of the paired HSI--TeX
        samples.
    \end{minipage}
\end{table*}

\subsection{Dataset overview}
\label{sec:dataset_overview}

\emph{TeX-1500} is a paired learning benchmark for LWIR TeX decomposition. It contains 1,522 real-scene samples, each pairing a calibrated wavelength-resolved thermal radiance cube with aligned temperature $T$, emissivity $e$, and texture $X$ supervision. The dataset is designed to move TeX decomposition from model-based, hand-tuned scene-specific optimization toward supervised HSI-to-TeX learning across scenes, wavelength layouts, and sensors.

Table~\ref{tab:dataset_summary} summarizes the two complementary data sources. The DARPA IH pushbroom subset~\cite{YellinConcurrentBandSelection2024} provides the large-scale outdoor backbone, with geographically separated sites, natural and man-made content, held-out validation/test locations, broad acquisition-time variation, and dense LWIR spectral sampling. Our FTIR subset provides the cross-sensor and close-range material counterpart, including plants, plaster, metal, face, cement, plastic, and glass under a different spectral layout. Figs.~\ref{fig:tex1650_train_sidewinder_samples}--\ref{fig:tex1650_ftir_test_wuhan_samples} show representative paired HSI--TeX samples from these splits.

\begin{figure}[!t]
    \centering
    \includegraphics[width=\columnwidth]{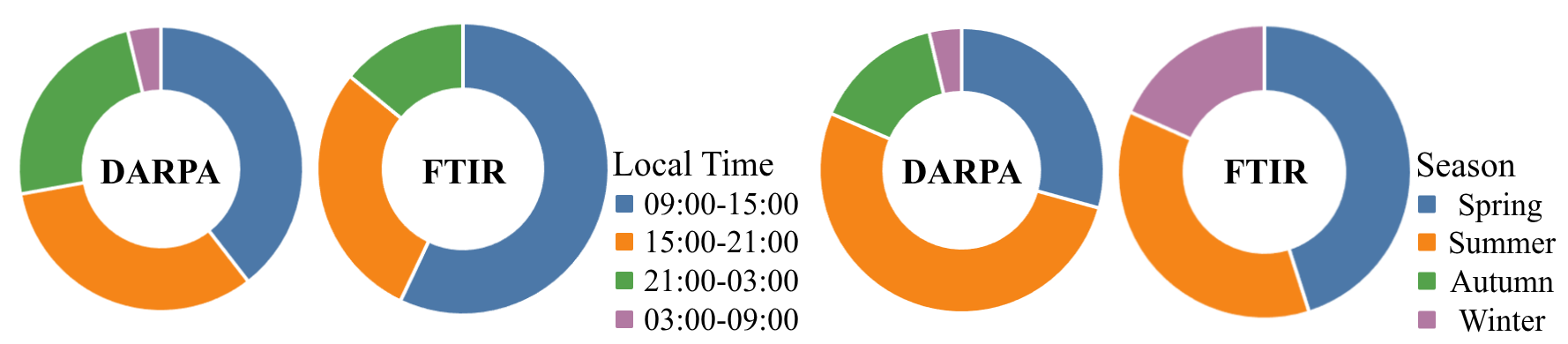}
    \caption{\textbf{Acquisition-time distribution of \emph{TeX-1500}.}
        Samples span diverse daytime, nighttime, and transitional thermal
        conditions, capturing variations in object heat state, thermal contrast,
        and downwelling environmental radiance.}
    \label{fig:dataset_time_distribution}
\end{figure}

\begin{figure}[!t]
    \centering
    \includegraphics[width=\columnwidth]{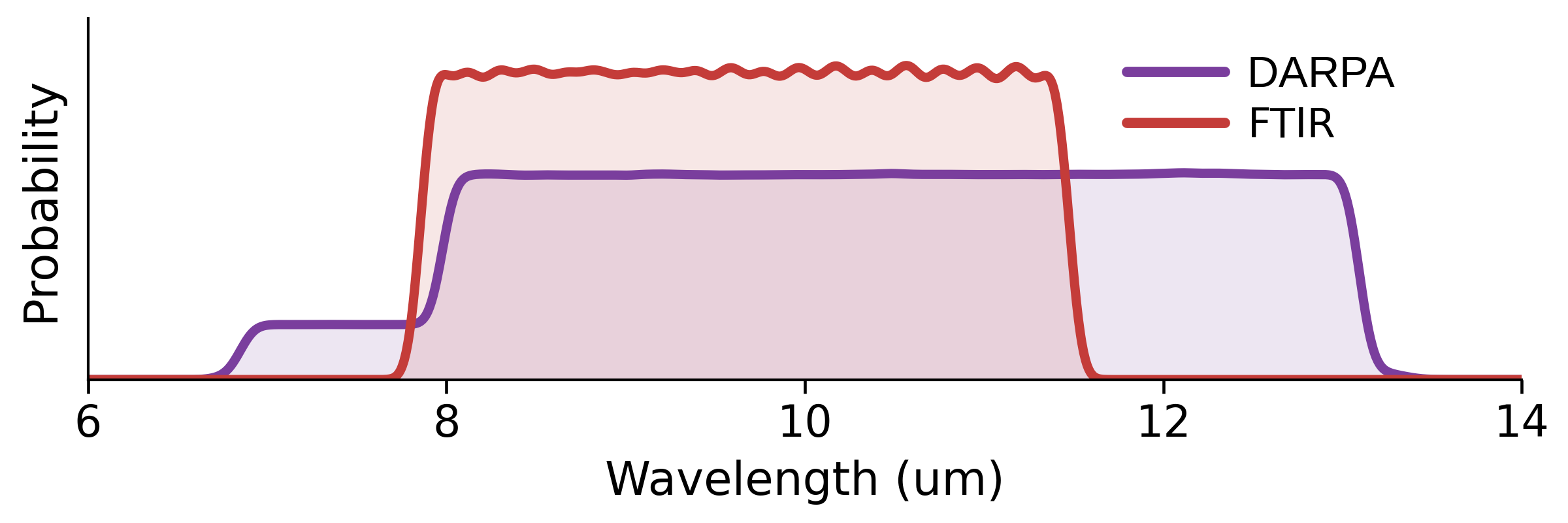}
    \caption{\textbf{Spectral coverage of \emph{TeX-1500}.}
        DARPA IH pushbroom and FTIR observations occupy thermal-infrared
        wavelength ranges with overlapping LWIR atmospheric-window coverage,
        while retaining sensor-specific band limits, sampling densities, and
        valid-band layouts.}
    \label{fig:dataset_wavelength_distribution}
\end{figure}

\begin{figure}[!t]
    \centering
    \includegraphics[width=\columnwidth]{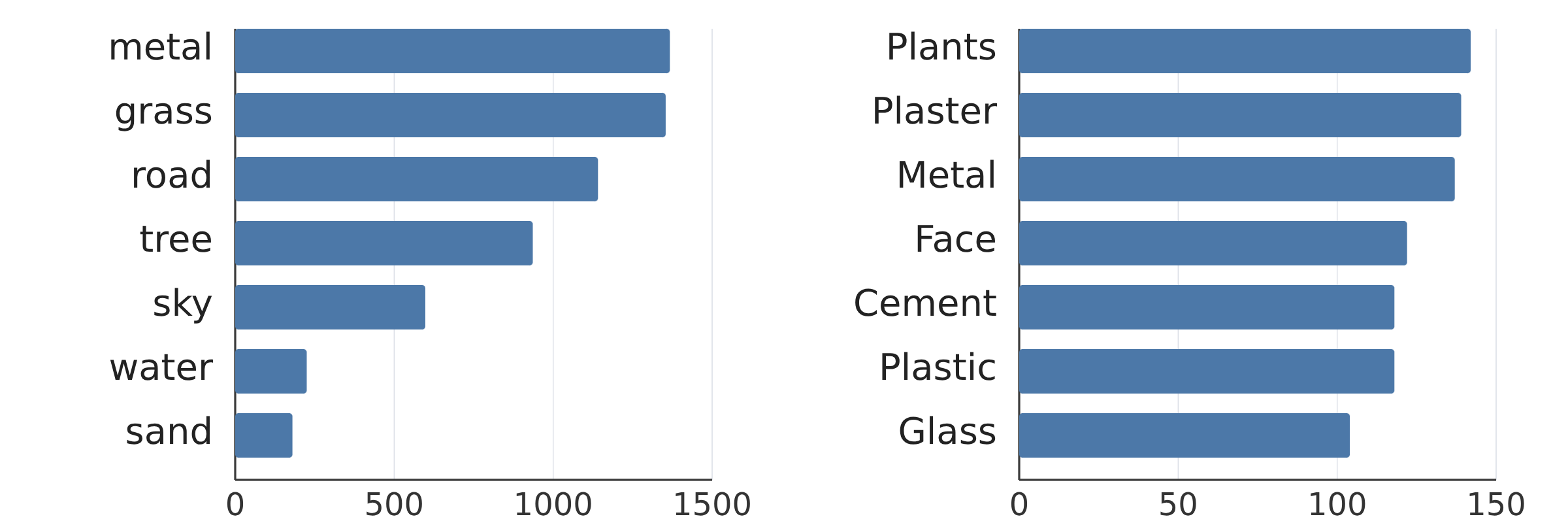}
    \caption{\textbf{Semantic-class distribution of \emph{TeX-1500}.}
        The DARPA IH subset (left) and FTIR subset (right) are summarized by
        their top-7 category proportions.}
    \label{fig:dataset_class_distribution}
\end{figure}

Figs.~\ref{fig:dataset_time_distribution}--\ref{fig:dataset_class_distribution} summarize the main distributional axes that a learning-based TeX decomposer must handle. Acquisition time changes object heat state, thermal contrast, and downwelling environmental radiance; spectral coverage determines which wavelength-dependent material cues are observable; and semantic composition controls the diversity of material and geometry encountered during training and evaluation. Together, these statistics define \emph{TeX-1500} as both a paired supervision resource and an evaluation setting for cross-scene, cross-band, and cross-sensor TeX decomposition.

\subsection{Dataset construction and distribution}
\label{sec:dataset_construction}

\emph{TeX-1500} converts DARPA IH outdoor pushbroom HSIs~\cite{YellinConcurrentBandSelection2024}\footnote{The original DARPA IH release contains 3,604 hyperspectral images; the current benchmark retains 1,380 DARPA IH images after repeated-scene removal for better class balance.}
and our FTIR acquisitions into paired HSI--TeX samples through a common
construction pipeline. The DARPA IH
train/validation/test split uses held-out scenes and acquisition dates to test
geographic and temporal generalization, while the FTIR split emphasizes changes
in sensor layout, scene content, and material type for cross-camera evaluation.

Each raw HSI is restored before TeX annotation. We remove corrupted bands,
suppress tractable degradations, calibrate wavelength shifts, and estimate the
sky signal required by the TeX construction protocol. The output is a calibrated
valid-band radiance cube paired with aligned temperature $T$, emissivity $e$,
and texture $X$ fields.

\subsubsection{HSI denoising}
For a degraded input $\mathcal{Y}\in\mathbb{R}^{H\times W\times C}$, distorted
boundary pixels are first cropped to obtain
$\mathcal{Y}'\in\mathbb{R}^{h\times w\times C}$. The band-wise stochastic
noise score $s_{1,k}$ is estimated via HySime~\cite{HySime_TGRS_2008}, with
threshold $\tau_1=0.01$. For pushbroom data, a stripe score is computed from
the cross-track row-mean profile $\mathbf r_k\in\mathbb{R}^h$ as
$s_{2,k}=\sqrt{h^{-1}\sum_{i=1}^{h}(r_{i,k}-\tilde r_{i,k})^2}$, where
$\tilde{\mathbf r}_k=\mathbf r_k\ast g_\sigma$ is a Gaussian-smoothed baseline
with $\sigma=10.0$. FTIR data use only $s_{1,k}$, whereas pushbroom data use
both degradation scores:
\begin{equation}
    \Omega_{\rm c}
    =
    \left\{
    k\in\{1,\ldots,C\}\;\middle|\;
    \begin{cases}
        s_{1,k}>\tau_1,                    & \text{FTIR}, \\
        s_{1,k}>\tau_1\lor s_{2,k}>\tau_2, & \text{PB}.
    \end{cases}
    \right\}.
    \label{eq:tex_candidate_band_set}
\end{equation}
Here the stripe threshold is $\tau_2=0.03$. To avoid over-masking, the
discarded ratio is capped by $\gamma_{\max}=0.7$. Let
$\mathcal T(\Omega,s,K)$ return the $K$ largest-score indices, with
$s_k=s_{1,k}$ for FTIR and $s_k=s_{1,k}+s_{2,k}$ for pushbroom. The dead-band
set is
\begin{equation}
    \Omega_{\rm d}
    =
    \begin{cases}
        \Omega_{\rm c}, & |\Omega_{\rm c}|\le \lfloor\gamma_{\max}C\rfloor, \\
        \mathcal T(\Omega_{\rm c},\{s_k\},\lfloor\gamma_{\max}C\rfloor),
                        & \text{otherwise}.
    \end{cases}
    \label{eq:tex_dead_band_set}
\end{equation}
The valid-band set is $\Omega_g=\{1,\ldots,C\}\setminus\Omega_{\rm d}$, and
the usable radiance cube is cropped as
\begin{equation}
    \mathcal{Y}_g
    =
    \mathcal{Y}'(:,:, \Omega_g)
    \in
    \mathbb{R}^{h\times w\times C_g},
    \qquad
    C_g=|\Omega_g|.
    \label{eq:tex_valid_band_hsi}
\end{equation}
On $\Omega_g$, pushbroom stripe artifacts are removed by decomposing the valid
cube into a clean component $\mathcal Z$ and stripe $\mathcal S$:
\begin{equation}
    \begin{aligned}
        \min_{\mathcal Z,\mathcal S}\quad
         &
        \frac{1}{2}\|\mathcal Y_g-\mathcal Z-\mathcal S\|_F^2
        +\lambda_1\|\nabla_x\mathcal Z\|_1
        +\alpha(s_{2,k})\|\nabla_y\mathcal Z\|_1
        \\
         & \quad
        +\lambda_3\|\nabla_{yy}\mathcal Z\|_1
        +\lambda_4\|\nabla_x\mathcal S\|_1
        +\lambda_5\|\mathcal S\|_1 ,
    \end{aligned}
    \label{eq:tex_destriping_objective}
\end{equation}
where $\alpha(s_{2,k})=m s_{2,k}$ with $m=2$. Residual stochastic
perturbations are removed via FastHyDe~\cite{zhuang2018fast} with $s_{1,k}$.

\subsubsection{HSI calibration}
We follow the first two stages of HAIR~\cite{dai2026hadarbasedthermalinfraredhyperspectral}
to obtain denoised HSIs, calibrated wavelengths, and calibrated sky signals.
Because view-specific sky simulation is impractical for each observation, we
combine a physics-generated atmospheric reference with the spectral features
measured in the HSI. First, the atmospheric record nearest in space and time is
used to generate a high-resolution near-ground downwelling reference
$\mathbf{s}_{\rm r} \in \mathbb{R}^{C_{\rm r}}$ with
libRadtran~\cite{libradtran_software_emde2016libradtran}. Second, the observed
atmospheric signature is extracted from the cleaned valid-band HSI.
Aligning these two spectra yields both the calibrated operating wavelengths and
the local calibrated sky signal.

Specifically, we average the cleaned valid-band cube
$\mathcal{Y}_{\rm denoised}\in\mathbb{R}^{h\times w\times C_g}$ over all
pixels to obtain a scene-level spectrum $\mathbf y\in\mathbb{R}^{C_g}$. Given
$\mathbf y$, asymmetric least-squares (ALS) baseline
separation~\cite{ALS_2025} decomposes it into a smooth thermal baseline
$\mathbf b$ and an atmospheric component by solving
\begin{equation}
    \min_{\mathbf b}
    \sum_{i=1}^{|\Omega_g|}
    \omega_i (y_i-b_i)^2
    +
    \beta
    \sum_{i=3}^{|\Omega_g|}
    (\Delta^2 b_i)^2 ,
    \label{eq:tex_als_baseline}
\end{equation}
where $\Delta^2b_i=b_i-2b_{i-1}+b_{i-2}$, $\beta=10^4$, and $\omega_i$ are
the iteratively updated asymmetric weights. The observed atmospheric signature
is $\mathbf{s}=\mathbf y-\mathbf b\in\mathbb{R}^{C_g}$.

Calibration distinguishes the nominal wavelength $\lambda_k$ from the actual
operating wavelength $\lambda_k^*=ak^2+bk+d$, with shift
$\Delta\lambda_k=\lambda_k^*-\lambda_k$. Given $\lambda_k^*$, the libRadtran
reference is projected to the sensor domain with a normalized Gaussian spectral
response function and matched to the observed amplitude via Z-score
normalization:
\begin{equation}
    \begin{aligned}
        \hat{\mathbf{s}}(k)
         & =
        \mu_{\mathbf{s}}
        +
        \sigma_{\mathbf{s}}
        \mathcal{Z}\!\left(
        \sum_{j=1}^{C_{\rm r}}
        \mathbf{s}_{\rm r}(j)
        \frac{
            \exp\!\left[-\frac{(\lambda_{{\rm r},j}-\lambda_k^*)^2}{2\sigma^2}\right]
        }{
            \sum_{\ell=1}^{C_{\rm r}}
            \exp\!\left[-\frac{(\lambda_{{\rm r},\ell}-\lambda_k^*)^2}{2\sigma^2}\right]
        }.\right)
    \end{aligned}
    \label{eq:tex_calibrated_sky_signal}
\end{equation}
Here $\mathcal{Z}(\cdot)$ denotes Z-score normalization, $\mu_{\mathbf{s}}$ and
$\sigma_{\mathbf{s}}$ are the mean and standard deviation of $\mathbf{s}$, and
$\sigma$ is the effective bandwidth. Unlike HAIR, which interpolates back to a
nominal wavelength grid, we compute Eq.~\eqref{eq:tex_calibrated_sky_signal}
only for $k\in\Omega_g$. The parameters are estimated by
\begin{equation}
    (\sigma^*,a^*,b^*,d^*)
    =
    \arg\min_{\sigma,a,b,d}
    \|\hat{\mathbf{s}}-\mathbf{s}\|_2^2 .
    \label{eq:tex_spectral_correction}
\end{equation}
The resulting $\hat{\mathbf{s}}(k)$ is the calibrated sky signal used by the
TeX label-construction protocol at wavelength $\nu=\lambda_k^*$. The calibrated
HSI is obtained by re-associating each valid radiance band with its calibrated
wavelength,
\begin{equation}
    \mathcal{Y}_{\rm c}(x,y,\lambda_k^*)
    =
    \mathcal{Y}_{\rm denoised}(x,y,k),
    \qquad
    k\in\Omega_g.
    \label{eq:tex_calibrated_hsi}
\end{equation}

\subsubsection{HSI--TeX pair construction}
\emph{TeX-1500} constructs paired supervision from the calibrated valid-band HSIs. The
TeX decomposition is implemented with an optimization pipeline
adapted from HADAR-SLOT~\cite{xu2026universalcomputationalthermalimaging}. We first
separate sky and non-sky pixels via matched filter, and then generate temperature,
emissivity, and texture fields under the same
construction protocol.\footnote{The physical derivation of the radiative-transfer
    model, including the new definition of texture $X$, will be presented in coming
    work (HADAR-v2). In this definition, $X$ is designed to be governed by scene
    geometry and scene emissivity, i.e., intrinsic object/scene properties, making
    it a stable and learnable target for HSI-to-TeX decomposition.}
Each dataset sample is stored as:
\begin{equation}
    \left(
    \mathcal{Y}_{\rm c}\in\mathbb{R}^{h\times w\times C_g},
    \{\lambda_k^*\}_{k\in\Omega_g},
    T,
    e(\{\lambda_k^*\}_{k\in\Omega_g}),
    X
    \right).
    \label{eq:tex_good_band_pair}
\end{equation}
This preserves the physical band locations, the image-specific valid-band
layout, and wavelength-independent TeX fields for paired HSI--TeX learning. We
do not interpolate the calibrated wavelengths back to the nominal grid, as in
HAIR, because this step does not add information useful for TeX decomposition.

\begin{algorithm}[H]
    \caption{HSI preprocess}
    \label{alg:tex_real_world_preprocessing}
    \small
    \begin{algorithmic}[1]
        \REQUIRE Raw HSI $\mathcal{Y}\in\mathbb{R}^{H\times W\times C}$, camera type
        $\xi\in\{\mathrm{PB},\mathrm{FTIR}\}$, nominal wavelengths $\{\lambda_k\}_{k=1}^C$,
        and local atmospheric record.
        \ENSURE Denoised HSI $\mathcal{Y}_{\rm denoised}$, calibrated wavelengths
        $\{\lambda_k^*\}_{k\in\Omega_g}$, and calibrated sky signal
        $\hat{\mathbf{s}}$.
        \STATE Crop distorted boundary pixels to obtain
        $\mathcal{Y}'\in\mathbb{R}^{h\times w\times C}$.
        \STATE Estimate stochastic noise scores $\{s_{1,k}\}_{k=1}^C$.
        \IF{$\xi=\mathrm{PB}$}
        \STATE Estimate stripe scores $\{s_{2,k}\}_{k=1}^C$.
        \ENDIF
        \STATE Determine the dead-band set $\Omega_{\rm d}$ and the valid-band set $\Omega_g$.
        \STATE Crop out the valid-band HSI $\mathcal{Y}_g=\mathcal{Y}'(:,:, \Omega_g)$.
        \IF{$\xi=\mathrm{PB}$}
        \STATE Remove stripe artifacts on $\Omega_g$.
        \ENDIF
        \STATE Denoise stochastic perturbations to obtain
        $\mathcal{Y}_{\rm denoised}$.
        \STATE Generate the reference $\mathbf{s}_{\rm r}$ from the nearest atmospheric record.
        \STATE Extract an observed atmospheric signature from $\mathcal{Y}_{\rm denoised}$.
        \STATE Estimate $\{\lambda_k^*\}_{k\in\Omega_g}$ and $\sigma$ by aligning the
        observed signature with the libRadtran reference.
        \STATE Obtain the calibrated sky signal $\hat{\mathbf{s}}(k)$ at $\nu=\lambda_k^*$.
        \STATE Re-associate valid radiance bands with calibrated wavelengths.
        \STATE \textbf{return} $\mathcal{Y}_{\rm denoised}$,
        $\{\lambda_k^*\}_{k\in\Omega_g}$, and $\hat{\mathbf{s}}$.
    \end{algorithmic}
\end{algorithm}

\subsection{Dataset quality assessment and validation}
\label{sec:dataset_properties}

We assess \emph{TeX-1500} through visual label quality, scene-level stability, and
learnability. The evaluation asks whether the construction produces coherent
paired TeX labels across scenes and whether a simple neural baseline can learn
the HSI-to-TeX mapping.

\subsubsection{Better TeX quality than prior works}
\begin{figure}[!t]
    \centering
    \includegraphics[width=0.49\textwidth]{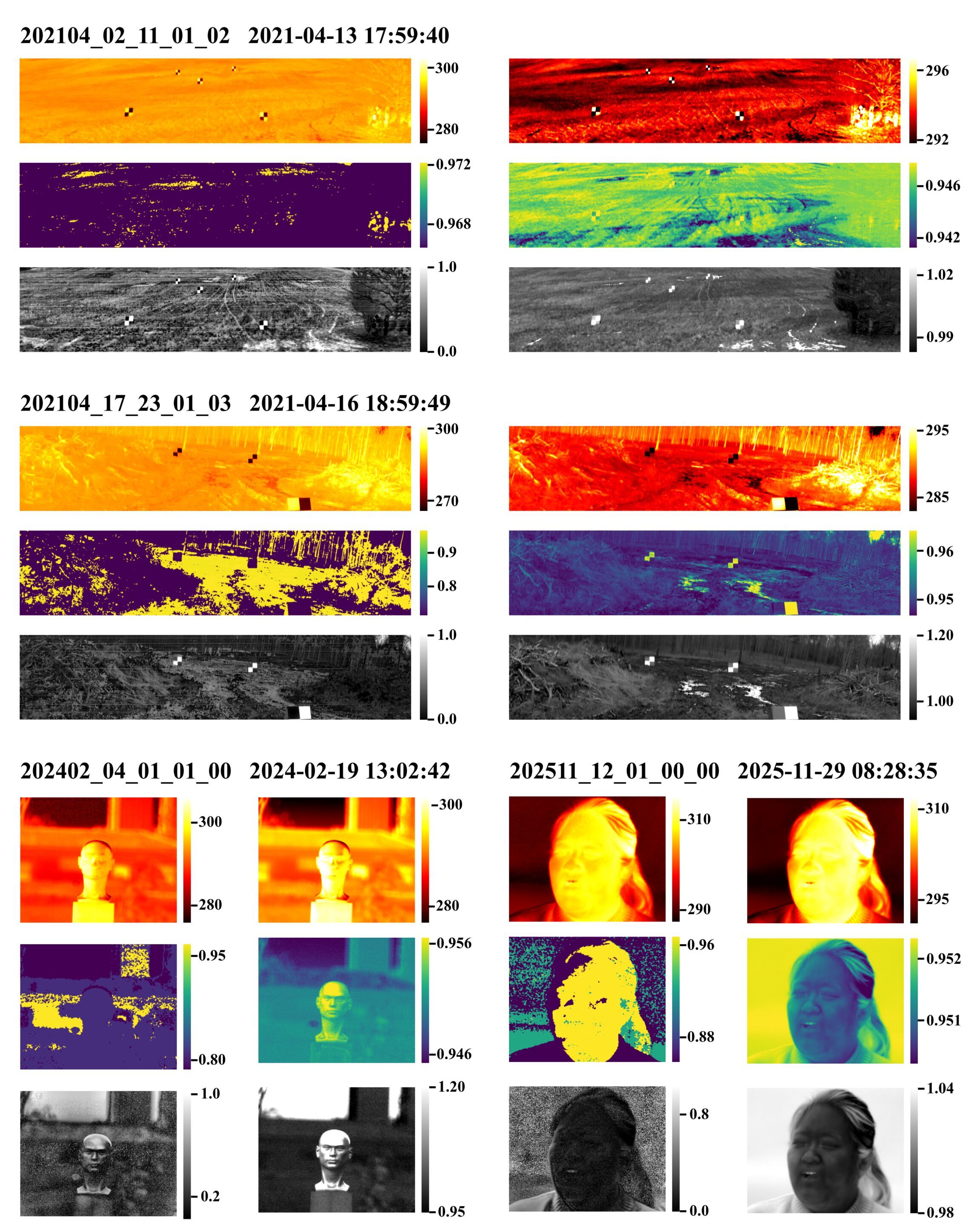}
    \caption{\textbf{TeX label (right) visual quality compared with HADAR-SGD (left).}}
    \label{fig:TeX_advantage_comparison}
\end{figure}

As shown in Fig.~\ref{fig:TeX_advantage_comparison}, the \emph{TeX-1500} construction
produces visually coherent TeX maps on both DARPA IH and FTIR scenes. Compared with
the TeX maps produced by HADAR-SGD~\cite{BaoHeatassistedDetectionRanging2023},
the \emph{TeX-1500} construction yields fields with smoother spatial
consistency, more balanced exposure, and clearer scene structure, while
HADAR-SGD outputs exhibit stronger spatial non-uniformity. This visual improvement is
important for learning-based TeX decomposition, where label artifacts can be
absorbed by the model as spurious supervision.

\subsubsection{Dataset stability and quality across scenes}
We verify construction stability by randomly sampling paired data from each
scene, as shown in
Figs.~\ref{fig:tex1650_train_sidewinder_samples}--\ref{fig:tex1650_ftir_test_wuhan_samples}.
These examples cover the DARPA IH training, validation, and testing splits as well
as the FTIR subset, exposing changes in location, season, sensor setting, and
scene composition. Across these subsets, the recovered TeX fields
remain visually stable under the same construction protocol: thermal ghosting
in the infrared observations is substantially reduced, while the temperature,
emissivity, and texture channels retain interpretable scene structures.

This scene-level stability is important because \emph{TeX-1500} is intended as a
benchmark for cross-camera and cross-environment TeX decomposition. These
samples show that the same construction pipeline can
produce coherent paired HSI--TeX data across outdoor pushbroom scenes and
close-range FTIR acquisitions. Quantitative evaluation of temperature and
emissivity consistency and accuracy is left to coming work.\footnote{Temperature
    and emissivity consistency and accuracy will be evaluated and presented in
    coming work (HADAR-v2).}

\subsection{Dataset learnability assessment}
\label{sec:dataset_learnability}

\begin{figure}[!t]
    \centering
    \includegraphics[width=0.49\textwidth]{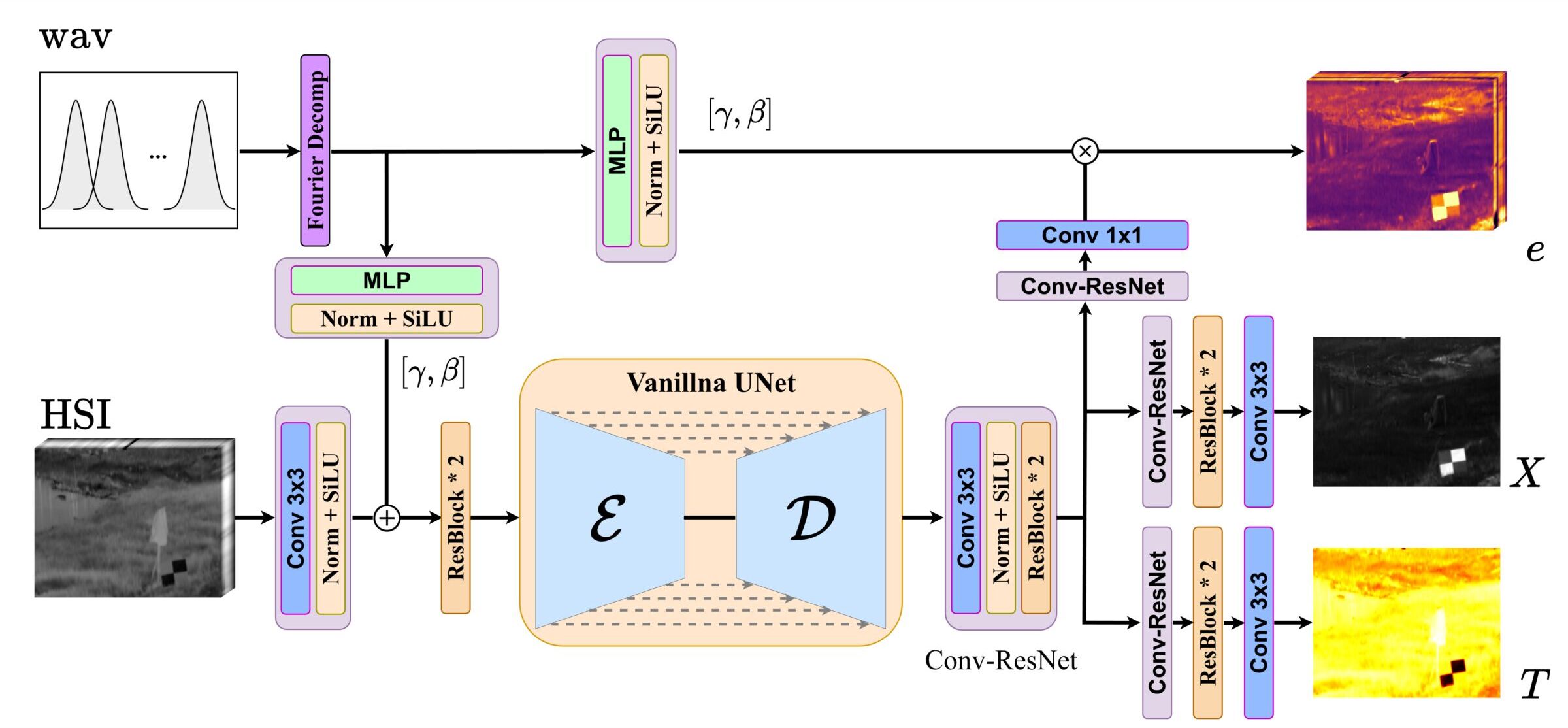}
    \caption{\textbf{TeX-UNet baseline for HSI-to-TeX inversion.}
        The model takes calibrated HSI bands and their wavelength positions as
        input and predicts temperature $T$, emissivity $e$, and texture $X$.}
    \label{fig:tex_unet_architecture}
\end{figure}

We further evaluate \emph{TeX-1500} as paired supervision for learning-based HSI-to-TeX
inversion. As an initial baseline, we train
TeX-UNet\footnote{Code and checkpoints are at \url{https://github.com/dccc2025/TeX-1500}.}
(Fig.~\ref{fig:tex_unet_architecture}), a lightweight vanilla U-Net-based
model that maps calibrated HSI bands and their spectra to temperature,
emissivity, and texture fields.

This experiment provides a direct learnability check: if a simple model trained
on \emph{TeX-1500} can reproduce held-out TeX maps and transfer to independent FTIR
measurements, then the paired labels provide usable supervision for subsequent
model development rather than only visually plausible reconstructions.

\subsubsection{Data preprocessing}
Before training, we clip each paired HSI--TeX sample with
\texttt{np.percentile(1,99)} to suppress isolated outliers. We normalize
emissivity $e$ and texture $X$, because $e$ is mainly used through spectral-line
shape for material recognition and $X$ is used for visual texture display.
Temperature $T$ is evaluated in Kelvin because its absolute physical accuracy
must be preserved.\footnote{The
    current framework does not solve low-reflectance-object recovery or physically
    accurate emissivity estimation, because physical emissivity recovery is affected
    by observation signal-to-noise ratio and radiometric correction. These issues,
    together with temperature and emissivity accuracy, will be presented, evaluated,
    and addressed in coming work (HADAR-v2).}\footnote{The \emph{TeX-1500} dataset based on HADAR-v2 will be
    released soon.}

\subsubsection{Training strategy}
We train TeX-UNet on the DARPA IH training split, use the DARPA IH validation split
for model selection, and report the DARPA IH test split in
Table~\ref{tab:tex_unet_inversion_results}. During training, each sample is
formed by a random $224\times224$ spatial crop and a random selection of 64 valid
spectral bands with their calibrated wavelength positions. Spatial rotations and
flips are used for augmentation, and spectral augmentation is applied through
random band sampling and small wavelength perturbations.

The training objective combines mean-squared reconstruction losses for the three
TeX fields with a third-order spectral smoothness regularizer on emissivity:
\begin{equation}
    \mathcal{L}
    =
    \|\hat{T}-T\|_2^2
    +
    \|\hat{e}-e\|_2^2
    +
    \|\hat{X}-X\|_2^2
    +
    \lambda_{\rm s}\|\Delta_{\lambda}^{3}\hat{e}\|_2^2 .
    \label{eq:tex_unet_loss}
\end{equation}
We set $\lambda_{\rm s}=0.01$. The remaining training hyperparameters are
summarized in Table~\ref{tab:tex_unet_training_settings}.

\subsubsection{Inference strategy}
During inference, we apply TeX-UNet to the full spatial image and handle
variable spectral layouts through repeated 64-band sampling, as detailed in
Algorithm~\ref{alg:tex_unet_full_scene_inference}. We sample until every valid
band is selected at least five times, average all TeX predictions, and evaluate
the resulting full-scene outputs on DARPA IH test scenes and FTIR zero-shot and
few-shot settings.

\begin{algorithm}[H]
    \caption{Full-scene variable-band TeX inference}
    \label{alg:tex_unet_full_scene_inference}
    \small
    \begin{algorithmic}[1]
        \REQUIRE Calibrated HSI $\mathcal{Y}_{\rm c}$, valid-band set
        $\Omega_g$, calibrated wavelengths $\{\lambda_k^*\}_{k\in\Omega_g}$,
        trained TeX-UNet $f_\theta$, band count $C_s=64$, minimum coverage
        $m=5$.
        \ENSURE Full-scene prediction $(\hat{T},\hat{e},\hat{X})$.
        \STATE Initialize prediction accumulator $\mathcal{A}\leftarrow 0$,
        count accumulator $\mathcal{N}\leftarrow 0$, and band coverage
        $n_k\leftarrow0$ for all $k\in\Omega_g$.
        \WHILE{$\min_{k\in\Omega_g} n_k < m$}
        \STATE Randomly sample $S\subset\Omega_g$ with $|S|=C_s$.
        \STATE Run $(T_S,e_S,X_S)=f_\theta(\mathcal{Y}_{\rm c}(:,:,S),
            \{\lambda_k^*\}_{k\in S})$.
        \STATE Add $(T_S,e_S,X_S)$ to $\mathcal{A}$ and increment
        $\mathcal{N}$.
        \STATE Update $n_k\leftarrow n_k+1$ for all $k\in S$.
        \ENDWHILE
        \STATE \textbf{return} $(\hat{T},\hat{e},\hat{X})=\mathcal{A}/\mathcal{N}$.
    \end{algorithmic}
\end{algorithm}

\begin{table}[H]
    \centering
    \caption{\textbf{TeX-UNet training hyperparameters.}}
    \label{tab:tex_unet_training_settings}
    \scriptsize
    \setlength{\tabcolsep}{4pt}
    \renewcommand{\arraystretch}{1.10}
    \begin{tabularx}{\columnwidth}{@{}p{2.15cm}>{\raggedright\arraybackslash}X@{}}
        \toprule
        Setting           & Value                                                                               \\
        \midrule
        Training input    & Random $224\times224$ crop and sampling
        of 64 valid bands                                                                                       \\
        Spectral encoding & Wavelength encoding enabled for the 64 selected bands                               \\
        Normalization     & Scene-level normalization with 1st/99th percentile clipping                         \\
        Optimizer         & AdamW, $\beta=(0.9,0.98)$, weight decay $10^{-4}$                                   \\
        DARPA IH training & Random init, 12k steps, LR $1.5\times10^{-4}$,
        1k warmup                                                                                               \\
        FTIR fine-tune    & FTIR from DARPA IH checkpoint, 2k steps,
        LR $3\times10^{-5}$, 200-step warmup                                                                    \\
        LR schedule       & Cosine decay to 0.05$\times$ the base LR                                            \\
        Batch/GPU         & 32 per GPU, bf16 precision, gradient clipping at 1.0                                \\
        U-Net config      & Depths $[2,2,2,2,2]$, channels
        $[64,128,256,352,480]$, trunk/head dims 240/224                                                         \\
        Inference         & Full-scene inference, detailed in Algorithm~\ref{alg:tex_unet_full_scene_inference} \\
        Hardware          & 8$\times$ NVIDIA RTX PRO 6000, 96 GB                                                \\
        \bottomrule
    \end{tabularx}
\end{table}

\begin{table}[H]
    \centering
    \caption{\textbf{TeX-UNet inversion results.}}
    \label{tab:tex_unet_inversion_results}
    \scriptsize
    \setlength{\tabcolsep}{4pt}
    \renewcommand{\arraystretch}{1.15}
    \resizebox{\columnwidth}{!}{%
        \begin{tabular}{@{}lcccccc@{}}
            \toprule
            \multirow{2}{*}{Test split}
             & \multicolumn{2}{c}{Temperature $T$}
             & \multicolumn{2}{c}{Emissivity $e$}
             & \multicolumn{2}{c}{Texture $X$}                 \\
            \cmidrule(lr){2-3}\cmidrule(lr){4-5}\cmidrule(l){6-7}
             & MAE (K)                             & MAPE (\%)
             & MSE                                 & SAM
             & MSE                                 & Deg       \\
            \midrule
            DARPA IH-test
             & 7.3284                              & 2.5488
             & 0.0453                              & 0.2267
             & 0.0311                              & 0.5206    \\
            FTIR-zeroshot-test
             & 5.8309                              & 1.9753
             & 0.0674                              & 0.0451
             & 0.0219                              & 0.2995    \\
            FTIR-fewshot-test
             & 4.1004                              & 1.3830
             & 0.0458                              & 0.1970
             & 0.0220                              & 0.2224    \\
            \bottomrule
        \end{tabular}
    }
    \vspace{4pt}
    \begin{minipage}{\columnwidth}
        \scriptsize
        \textit{Note.} $e$ and $X$ are normalized.
    \end{minipage}
\end{table}
\vspace{6pt}

\begin{figure}[!t]
    \centering
    \includegraphics[width=\columnwidth]{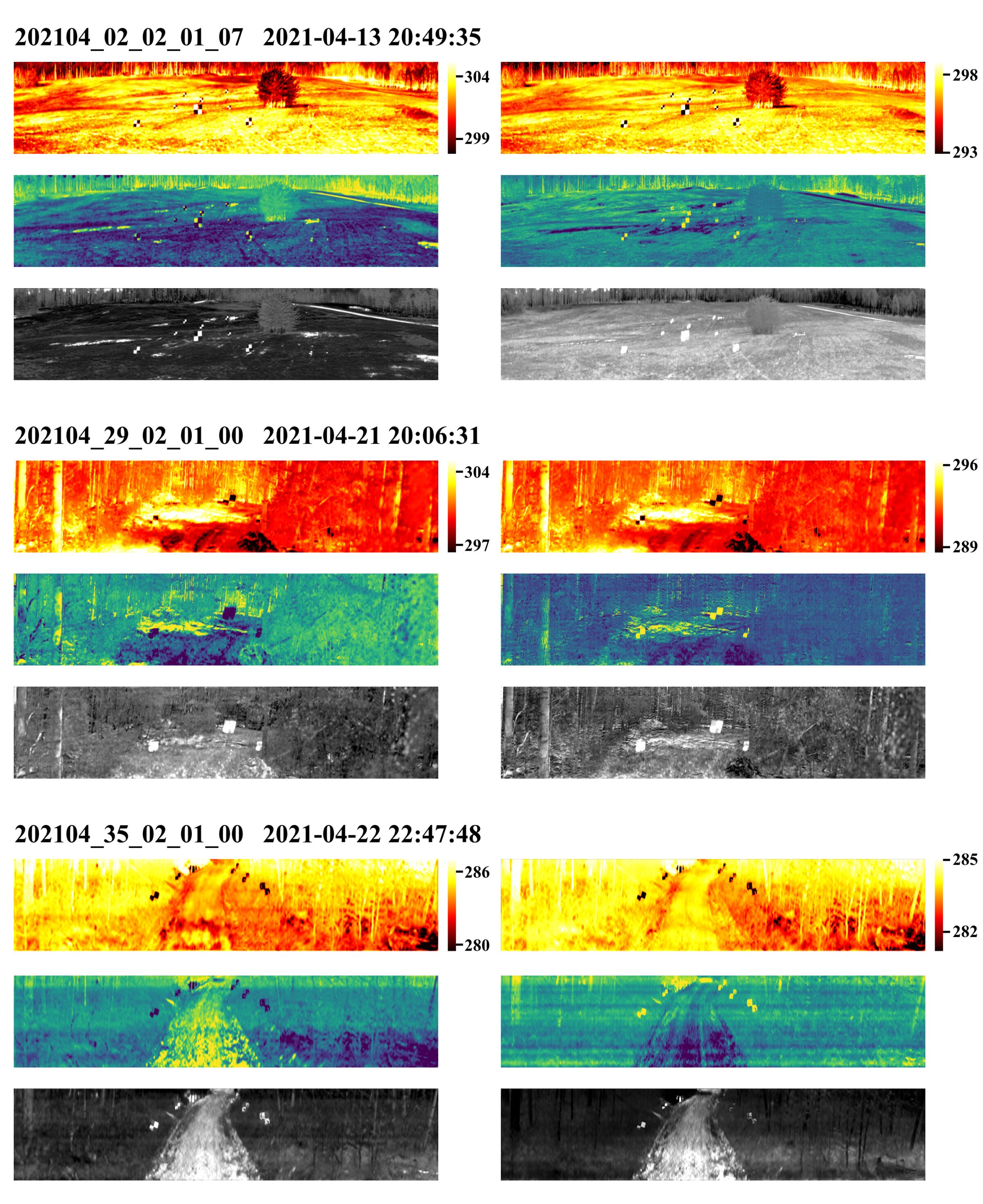}
    \caption{\textbf{TeX-UNet results on the DARPA IH test split.}
        The model is trained on the DARPA IH training split and evaluated on
        DARPA IH test scenes. The predicted TeX maps (left) preserve the main
        temperature, emissivity, and texture structures of the ground-truth
        labels (right).}
    \label{fig:tex_unet_darpa_qualitative}
\end{figure}

\begin{figure}[!t]
    \centering
    \includegraphics[width=\columnwidth]{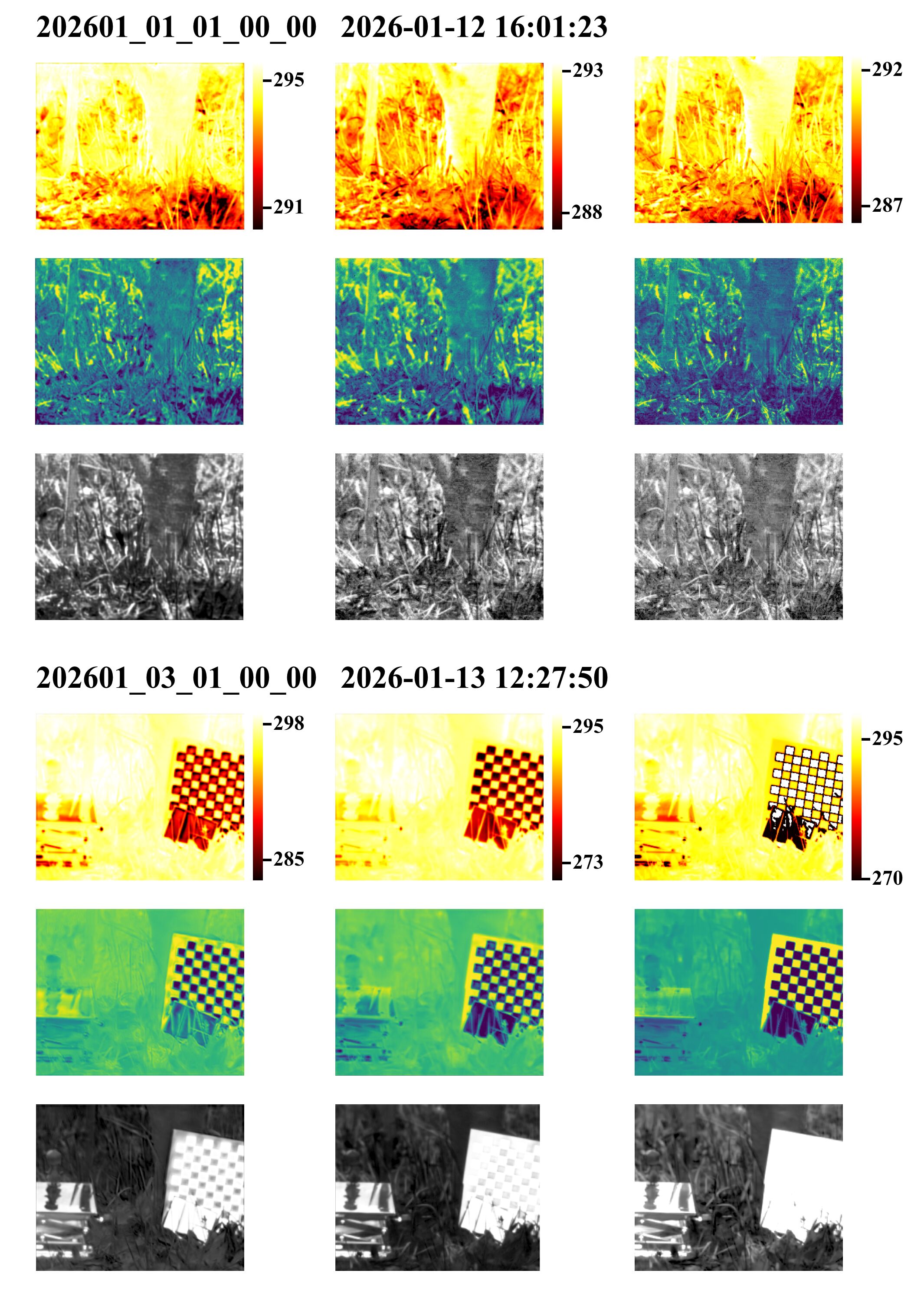}
    \caption{\textbf{Zero-shot and few-shot transfer to FTIR scenes.}
        TeX-UNet is first trained on the DARPA IH training split and then evaluated
        on FTIR data. The columns compare zero-shot prediction (left),
        few-shot fine-tuned prediction (middle) and ground truth (right), respectively,
        for $T$, $e$, and $X$; few-shot fine-tuning produces outputs closer to
        the ground truth than zero-shot transfer.}
    \label{fig:tex_unet_ftir_transfer_qualitative}
\end{figure}

The preliminary results show that TeX-UNet produces TeX maps close to the
ground-truth labels on DARPA IH test scenes
(Fig.~\ref{fig:tex_unet_darpa_qualitative}). Our FTIR scenes, zero-shot
transfer already recovers recognizable TeX structures, and few-shot fine-tuning
further moves the predictions toward the ground truth
(Fig.~\ref{fig:tex_unet_ftir_transfer_qualitative}). This gap between zero-shot
and few-shot performance indicates that FTIR HSI provides learnable
cross-sensor supervision rather than only a visualization target. Temperature is
reported in Kelvin, while emissivity and texture are reported in normalized
space (Table~\ref{tab:tex_unet_inversion_results}). Since emissivity amplitude
is sensitive to sensor-dependent calibration, SAM is included as the primary
spectral-shape metric for $e$.

\section{Conclusion}
This paper presents \emph{TeX-1500}, a paired LWIR HSI--TeX dataset and initial
benchmark for learning-based temperature--emissivity--texture decomposition. The
dataset contains 1,522 calibrated real-scene samples from DARPA IH pushbroom data
and FTIR acquisitions, pairing wavelength-resolved thermal radiance with
aligned temperature, emissivity, and texture fields. By combining diverse outdoor
pushbroom scenes with close-range FTIR measurements, \emph{TeX-1500} provides a data
foundation for studying cross-scene, cross-band, and cross-sensor thermal
perception grounded in physical quantities.

We further validate the dataset through visual quality analysis, scene-level
stability checks, and a TeX-UNet baseline. The results show that the constructed
TeX labels are visually coherent across DARPA IH and FTIR scenes, that TeX-UNet can
recover held-out DARPA IH TeX structures, and that FTIR zero-shot/few-shot
experiments provide a measurable cross-sensor learnability setting. \emph{TeX-1500}
therefore turns TeX decomposition from a primarily model-based, hand-tuned inverse
problem into a supervised benchmark for developing data-driven HSI-to-TeX
methods.

\section*{Acknowledgment}
\noindent
The authors thank Liqin Cao and Yanfei Zhong from the State Key Laboratory of
Information Engineering in Surveying, Mapping and Remote Sensing, Wuhan
University, for providing the imaging equipment and scene support. We thank
Xin Yuan from the School of Engineering, Westlake University, for his
guidance on dataset construction and planning. We also thank Du Wang, Chenjun
Zhao, and Jiashuo Chen for their assistance with data
collection.

\FloatBarrier
\clearpage
\raggedbottom
\begingroup
\setlength{\abovecaptionskip}{4pt}
\setlength{\belowcaptionskip}{0pt}
\newlength{\suppfigureheight}
\setlength{\suppfigureheight}{0.831\textheight}
\newcommand{\suppwidegraphic}[1]{%
    \parbox[c][0.86\textheight][c]{\textwidth}{%
        \centering
        \includegraphics[width=1.056\textwidth,height=\suppfigureheight,keepaspectratio]{#1}%
    }%
}
\newcommand{\suppcolgraphic}[1]{%
    \parbox[c][0.86\textheight][c]{\columnwidth}{%
        \centering
        \includegraphics[width=1.078\columnwidth,height=\suppfigureheight,keepaspectratio]{#1}%
    }%
}
\newcommand{\suppwidecaption}[4]{%
    \begin{minipage}{#1\textwidth}
        \caption[]{#3}
        \label{#4}
    \end{minipage}%
}
\newcommand{\suppcolcaption}[3]{%
    \begin{minipage}{0.88\columnwidth}
        \caption[]{#2}
        \label{#3}
    \end{minipage}%
}

\begin{figure*}[p]
    \centering
    \suppwidegraphic{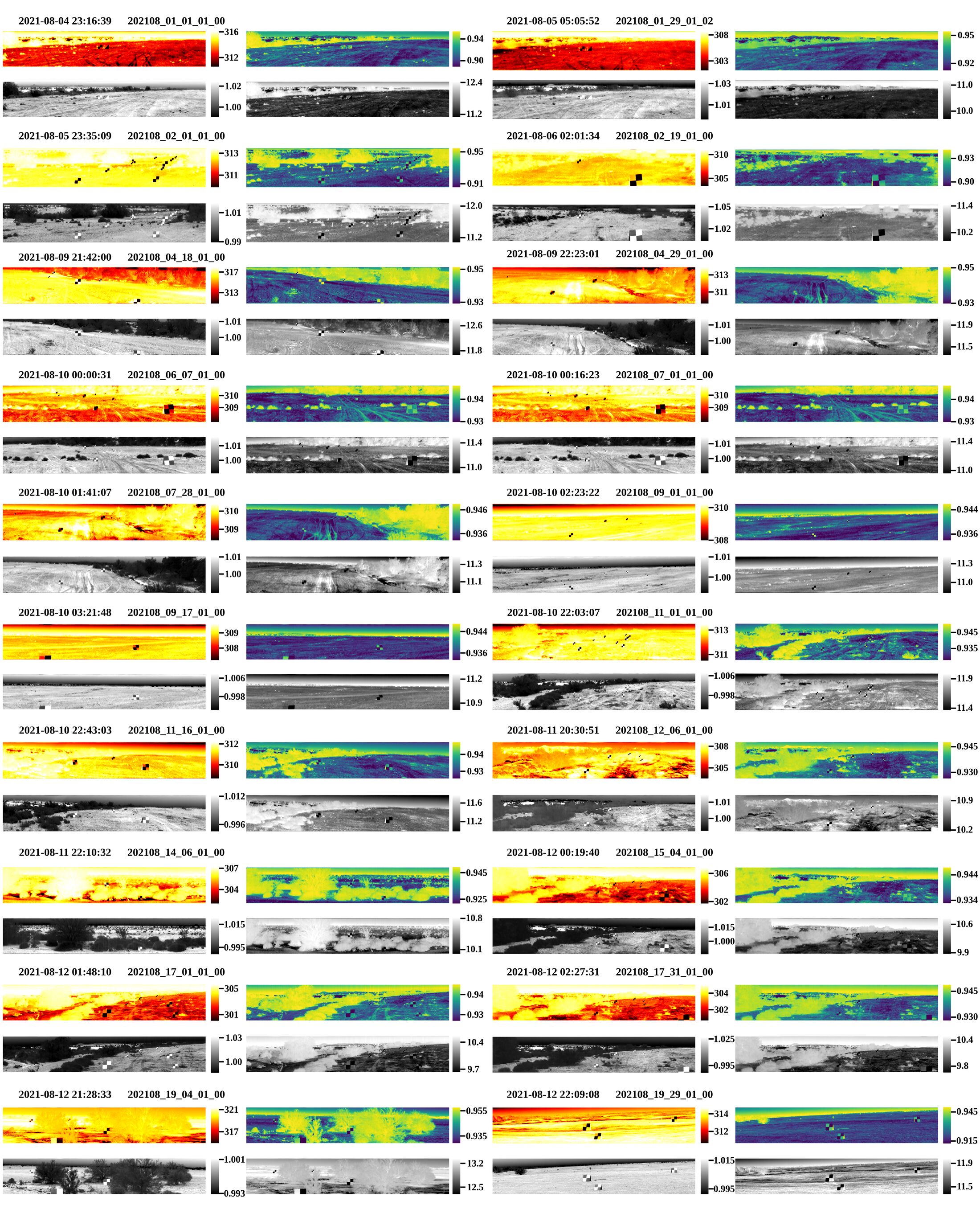}
    \suppwidecaption{0.91}{Training set samples from \emph{TeX-1500} DARPA IH pushbroom subset at Sidewinder Range}{\textbf{Training set samples from \emph{TeX-1500} DARPA IH pushbroom subset at Sidewinder Range, TPG, AZ in August 2021.}
        Panels show $T$ (in K), $e$, $X$, and HSI-band radiance ($\mathrm{W}\cdot \mathrm{m}^{-2}\cdot \mathrm{sr}^{-1}\cdot \mu\mathrm{m}^{-1}$).}{fig:tex1650_train_sidewinder_samples}
\end{figure*}

\begin{figure*}[p]
    \centering
    \suppwidegraphic{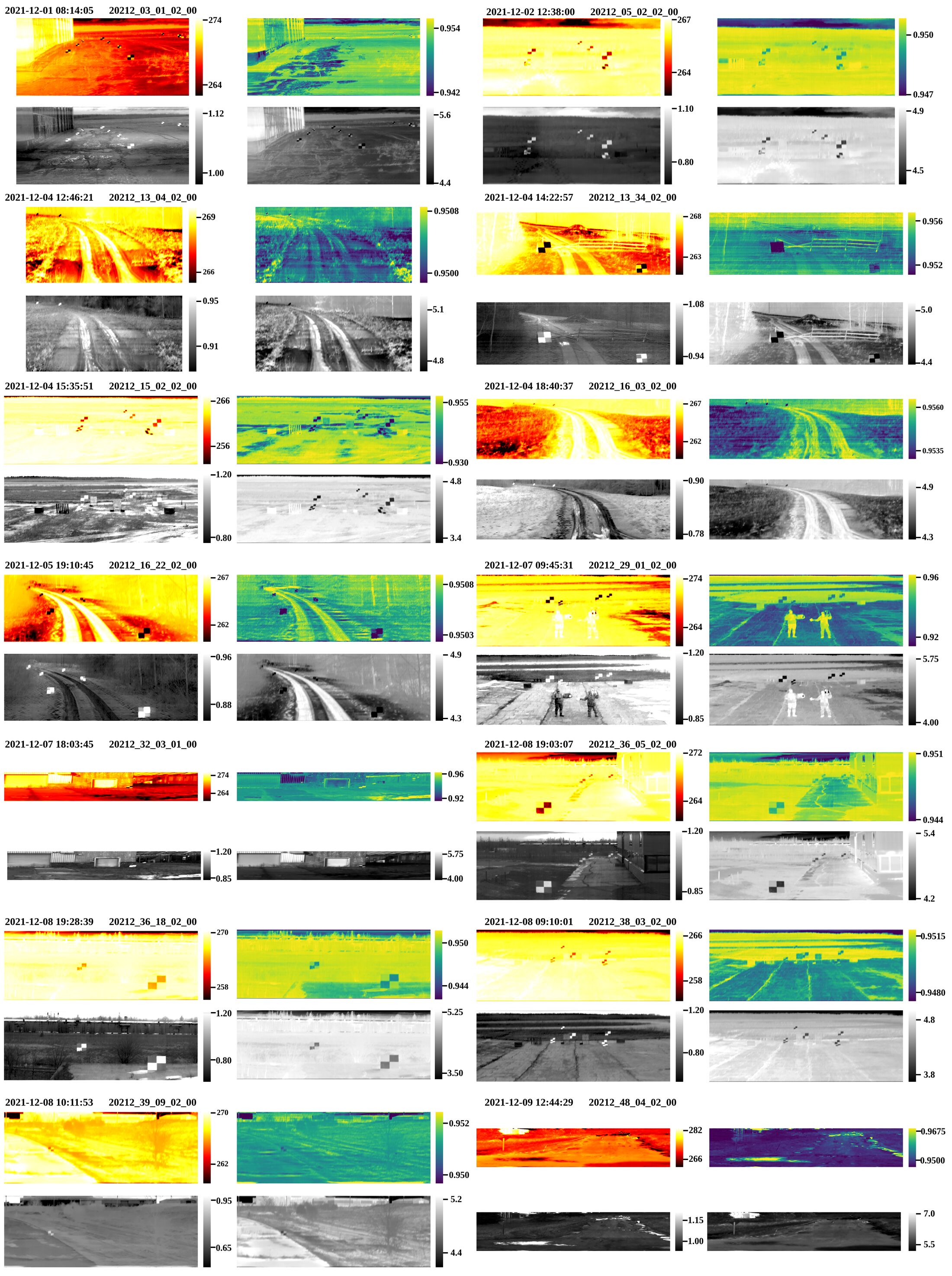}
    \suppwidecaption{0.86}{Training set samples from \emph{TeX-1500} DARPA IH pushbroom subset at Loring Commerce Center}{\textbf{Training set samples from \emph{TeX-1500} DARPA IH pushbroom subset at Loring Commerce Center, ME in December 2021.}
        Panels show $T$ (in K), $e$, $X$, and HSI-band radiance ($\mathrm{W}\cdot \mathrm{m}^{-2}\cdot \mathrm{sr}^{-1}\cdot \mu\mathrm{m}^{-1}$).}{fig:tex1650_train_loring_samples}
\end{figure*}

\begin{figure*}[p]
    \centering
    \suppwidegraphic{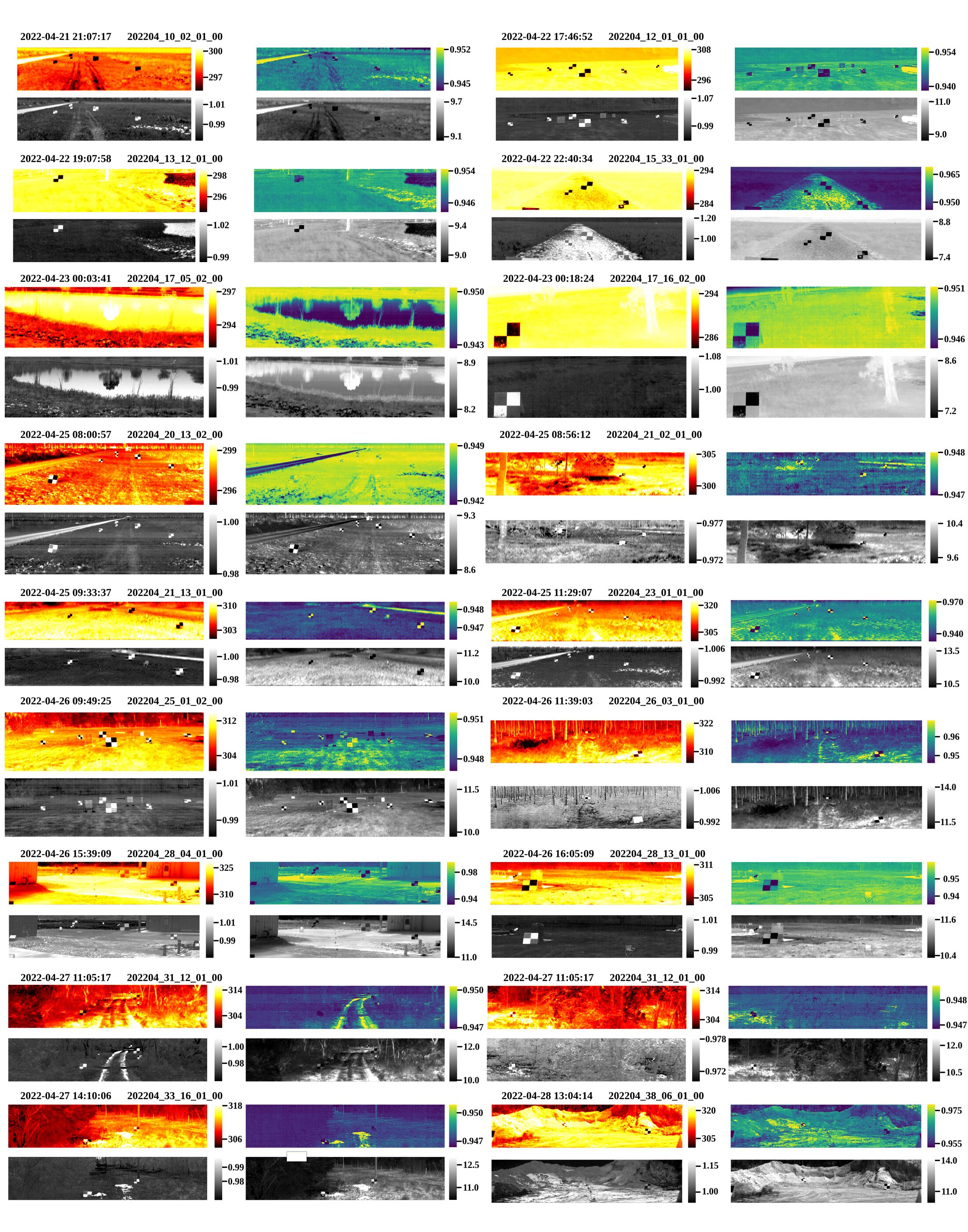}
    \suppwidecaption{0.88}{Training set samples from \emph{TeX-1500} DARPA IH pushbroom subset at Avon Park Air Force Range}{\textbf{Training set samples from \emph{TeX-1500} DARPA IH pushbroom subset at Avon Park Air Force Range, FL in April 2022.}
        Panels show $T$ (in K), $e$, $X$, and HSI-band radiance ($\mathrm{W}\cdot \mathrm{m}^{-2}\cdot \mathrm{sr}^{-1}\cdot \mu\mathrm{m}^{-1}$).}{fig:tex1650_train_avon_park_fl_samples}
\end{figure*}

\begin{figure*}[p]
    \centering
    \suppwidegraphic{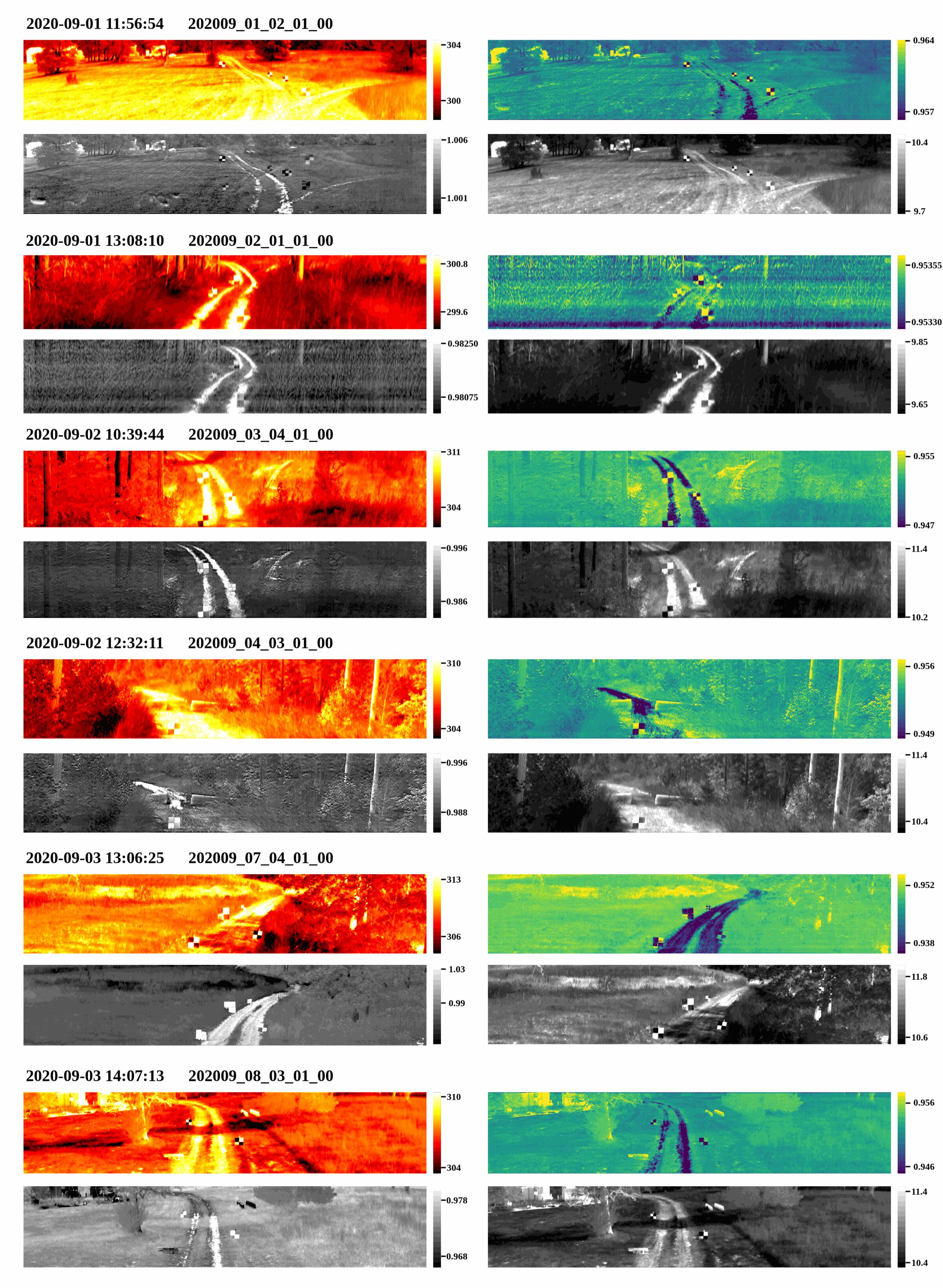}
    \suppwidecaption{0.86}{Validation set samples from \emph{TeX-1500} DARPA IH pushbroom subset at Sidewinder Range}{\textbf{Validation set samples from \emph{TeX-1500} DARPA IH pushbroom subsetat Sidewinder Range, TPG, AZ in September 2020.}
        Panels show $T$ (in K), $e$, $X$, and HSI-band radiance ($\mathrm{W}\cdot \mathrm{m}^{-2}\cdot \mathrm{sr}^{-1}\cdot \mu\mathrm{m}^{-1}$).}{fig:tex1650_valid_sidewinder_samples}
\end{figure*}

\begin{figure*}[p]
    \centering
    \suppwidegraphic{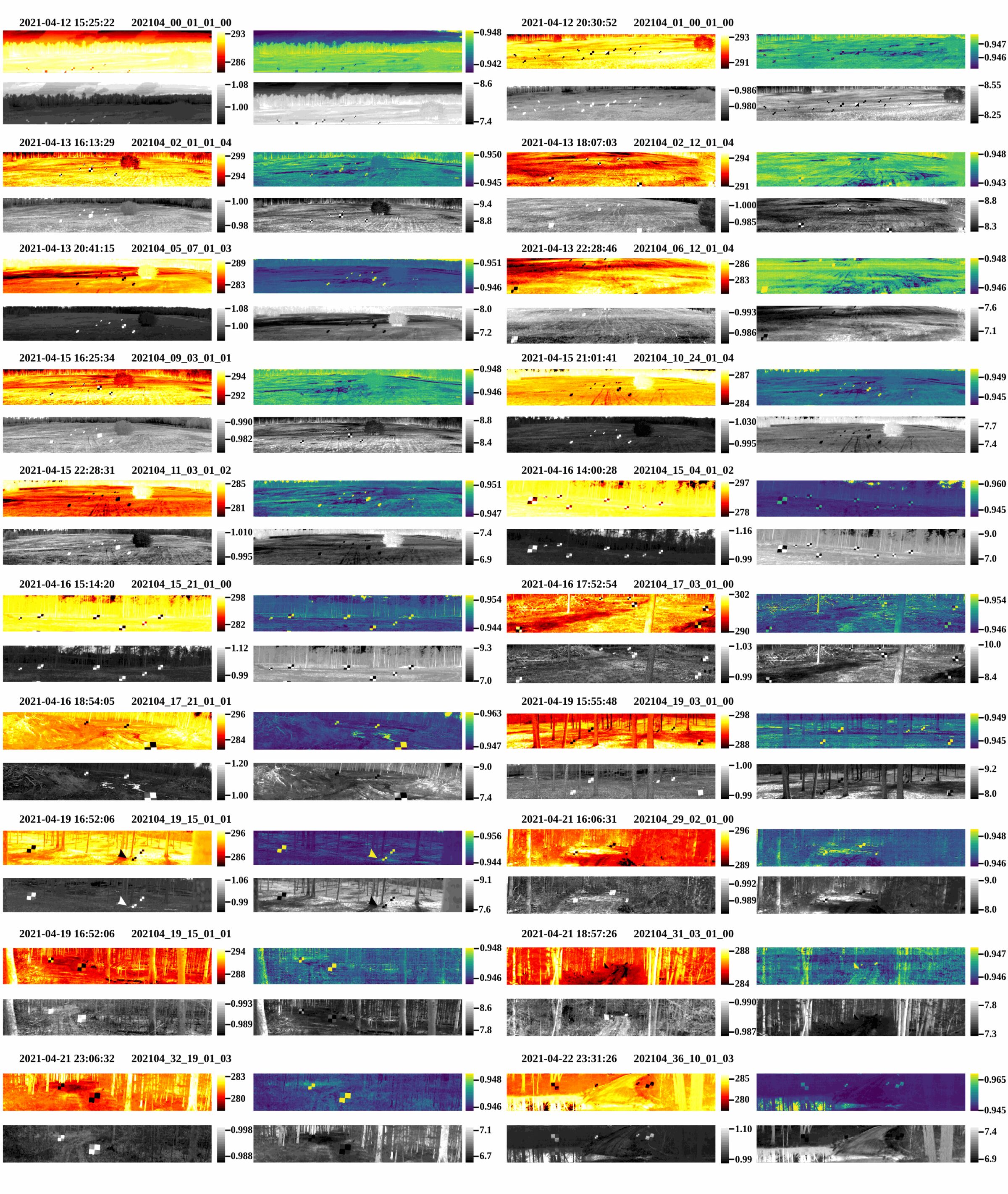}
    \suppwidecaption{0.94}{Testing set samples from \emph{TeX-1500} DARPA IH pushbroom subset at Fort A. P. Hill}{\textbf{Testing set samples from \emph{TeX-1500} DARPA IH pushbroom subset at Fort A. P. Hill, VA in April 2021.}
        Panels show $T$ (in K), $e$, $X$, and HSI-band radiance ($\mathrm{W}\cdot \mathrm{m}^{-2}\cdot \mathrm{sr}^{-1}\cdot \mu\mathrm{m}^{-1}$).}{fig:tex1650_test_fort_a_p_hill_samples}
\end{figure*}

\begin{figure}[p]
    \centering
    \suppcolgraphic{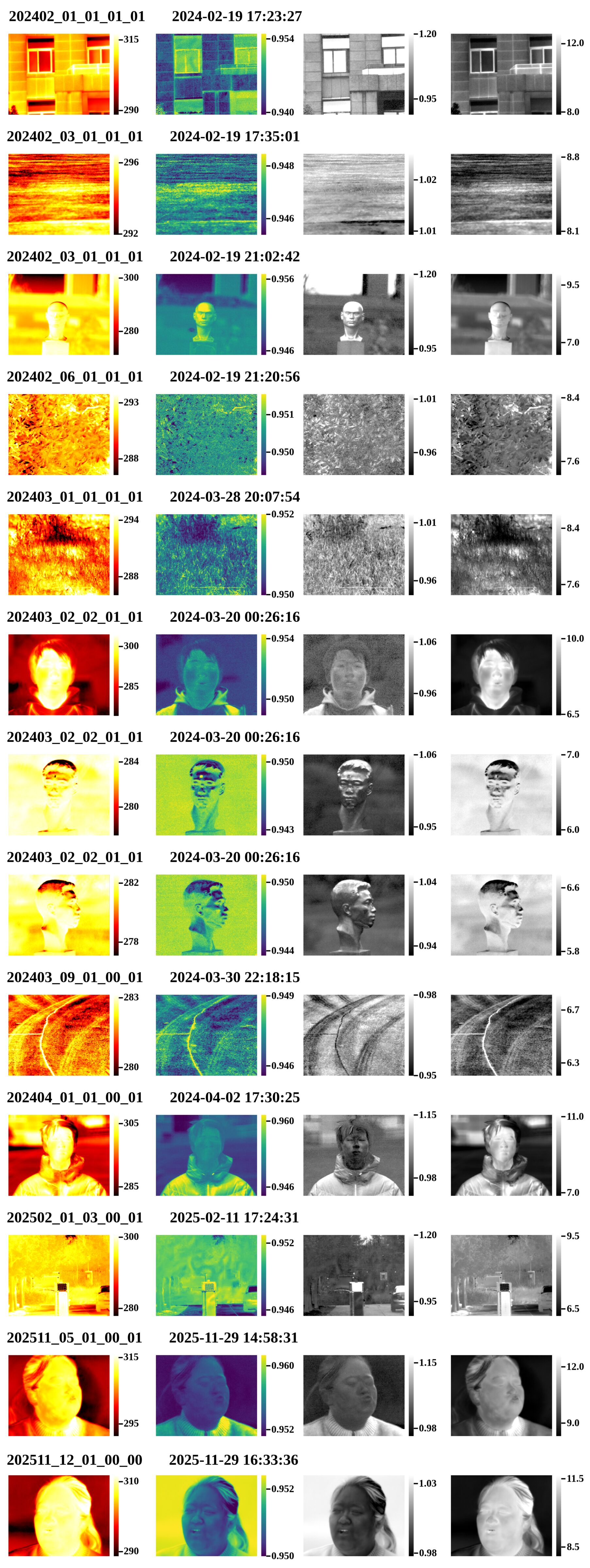}
    \suppcolcaption{Training set samples from \emph{TeX-1500} FTIR subset at Wuhan University}{\textbf{Training set samples from \emph{TeX-1500} FTIR subset.}
        Panels show $T$ (in K), $e$, $X$, and HSI-band radiance ($\mathrm{W}\cdot \mathrm{m}^{-2}\cdot \mathrm{sr}^{-1}\cdot \mu\mathrm{m}^{-1}$).}{fig:tex1650_ftir_train_wuhan_samples}
\end{figure}

\begin{figure}[p]
    \centering
    \suppcolgraphic{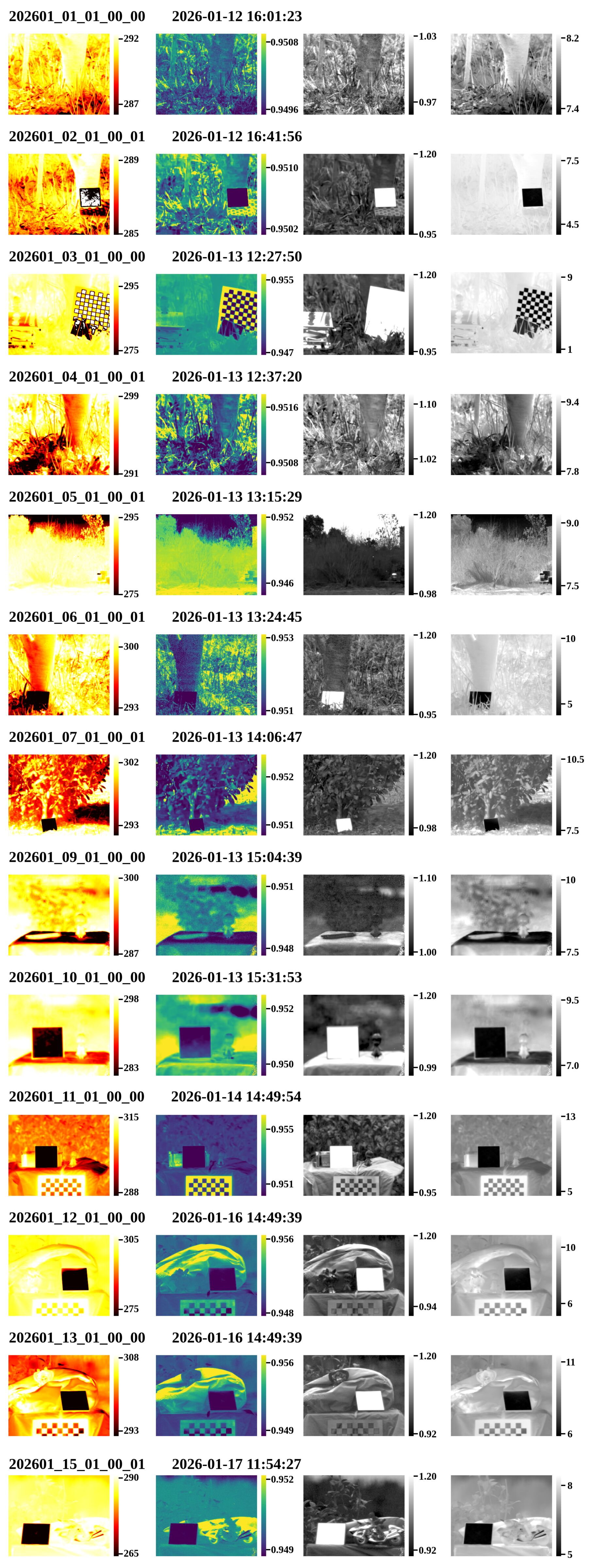}
    \suppcolcaption{Testing set samples from \emph{TeX-1500} FTIR subset at Wuhan University}{\textbf{Testing set samples from \emph{TeX-1500} FTIR subset.}
        Panels show $T$ (in K), $e$, $X$, and HSI-band radiance ($\mathrm{W}\cdot \mathrm{m}^{-2}\cdot \mathrm{sr}^{-1}\cdot \mu\mathrm{m}^{-1}$).}{fig:tex1650_ftir_test_wuhan_samples}
\end{figure}

\endgroup


\FloatBarrier
\clearpage

\bibliographystyle{IEEEtran}
\bibliography{references}

@article{cave_dataset_2010_tip,
  title={Generalized assorted pixel camera: postcapture control of resolution, dynamic range, and spectrum},
  author={Yasuma, Fumihito and Mitsunaga, Tomoo and Iso, Daisuke and Nayar, Shree K},
  journal={IEEE transactions on image processing},
  volume={19},
  number={9},
  pages={2241--2253},
  year={2010},
  publisher={IEEE}
}

@article{satmae_2022_nips,
  title={Satmae: Pre-training transformers for temporal and multi-spectral satellite imagery},
  author={Cong, Yezhen and Khanna, Samar and Meng, Chenlin and Liu, Patrick and Rozi, Erik and He, Yutong and Burke, Marshall and Lobell, David and Ermon, Stefano},
  journal={Advances in Neural Information Processing Systems},
  volume={35},
  pages={197--211},
  year={2022}
}

@ARTICLE{hyperSIGMA_2025_tpami,
  author={Wang, Di and Hu, Meiqi and Jin, Yao and Miao, Yuchun and Yang, Jiaqi and Xu, Yichu and Qin, Xiaolei and Ma, Jiaqi and Sun, Lingyu and Li, Chenxing and Fu, Chuan and Chen, Hongruixuan and Han, Chengxi and Yokoya, Naoto and Zhang, Jing and Xu, Minqiang and Liu, Lin and Zhang, Lefei and Wu, Chen and Du, Bo and Tao, Dacheng and Zhang, Liangpei},
  journal={IEEE Transactions on Pattern Analysis and Machine Intelligence}, 
  title={HyperSIGMA: Hyperspectral Intelligence Comprehension Foundation Model}, 
  year={2025},
  volume={47},
  number={8},
  pages={6427-6444},
  doi={10.1109/TPAMI.2025.3557581}}

@inproceedings{wangdu2025spectral_igarss,
  title={Spectral Noise Resistance Split Window Atmospheric Compensation for Airborne Thermal Infrared Hyperspectral},
  author={Wang, Du and Cao, Liqin and Gao, Lyuzhou and Ye, Fawang and Zhong, Yanfei},
  booktitle={IGARSS 2025-2025 IEEE International Geoscience and Remote Sensing Symposium},
  pages={1244--1248},
  year={2025},
  organization={IEEE}
}

@article{ieee_grss_datasets_2015_jstars,
  title={Processing of multiresolution thermal hyperspectral and digital color data: Outcome of the 2014 IEEE GRSS data fusion contest},
  author={Liao, Wenzhi and Huang, Xin and Van Coillie, Frieke and Gautama, Sidharta and Pi{\v{z}}urica, Aleksandra and Philips, Wilfried and Liu, Hui and Zhu, Tingting and Shimoni, Michal and Moser, Gabriele and others},
  journal={IEEE Journal of Selected Topics in Applied Earth Observations and Remote Sensing},
  volume={8},
  number={6},
  pages={2984--2996},
  year={2015},
  publisher={IEEE}
}

@article{spectralgpt_2024_tpami,
  title={SpectralGPT: Spectral remote sensing foundation model},
  author={Hong, Danfeng and Zhang, Bing and Li, Xuyang and Li, Yuxuan and Li, Chenyu and Yao, Jing and Yokoya, Naoto and Li, Hao and Ghamisi, Pedram and Jia, Xiuping and others},
  journal={IEEE transactions on pattern analysis and machine intelligence},
  volume={46},
  number={8},
  pages={5227--5244},
  year={2024},
  publisher={IEEE}
}

@inproceedings{boson_nighttime_2023_iros,
  title={Long-range uav thermal geo-localization with satellite imagery},
  author={Xiao, Jiuhong and Tortei, Daniel and Roura, Eloy and Loianno, Giuseppe},
  booktitle={2023 IEEE/RSJ International Conference on Intelligent Robots and Systems (IROS)},
  pages={5820--5827},
  year={2023},
  organization={IEEE}
}

@article{spectralearth_2025_jstars,
  title={Spectralearth: Training hyperspectral foundation models at scale},
  author={Braham, Nassim Ait Ali and Albrecht, Conrad M and Mairal, Julien and Chanussot, Jocelyn and Wang, Yi and Zhu, Xiao Xiang},
  journal={IEEE Journal of Selected Topics in Applied Earth Observations and Remote Sensing},
  year={2025},
  publisher={IEEE}
}

@article{thermalgen_2026_nips,
  title={Thermalgen: Style-disentangled flow-based generative models for rgb-to-thermal image translation},
  author={Xiao, Jiuhong and Nayak, Roshan and Zhang, Ning and Tortei, Daniel and Loianno, Giuseppe},
  journal={Advances in Neural Information Processing Systems},
  volume={38},
  pages={33111--33133},
  year={2026}
}

@article{mao2026pid,
  title={PID: Physics-informed diffusion model for infrared image generation},
  author={Mao, Fangyuan and Mei, Jilin and Lu, Shun and Liu, Fuyang and Chen, Liang and Zhao, Fangzhou and Hu, Yu},
  journal={Pattern Recognition},
  volume={169},
  pages={111816},
  year={2026},
  publisher={Elsevier}
}

@article{zhao2026thermal,
  title={Thermal-physics-informed 3D Gaussian Splatting for infrared images rendering},
  author={Zhao, Yunpeng and Sun, Hui and Liu, Chao and Qi, Zhenheng},
  journal={Infrared Physics \& Technology},
  pages={106593},
  year={2026},
  publisher={Elsevier}
}

@article{meng2025pcmamba,
  title={PCMamba: Physics-informed cross-modal state space model for dual-camera compressive hyperspectral imaging},
  author={Meng, Ge and Cai, Zhongnan and Tu, Jingyan and Wang, Yingying and Li, Chenxin and Huang, Yue and Ding, Xinghao},
  journal={arXiv preprint arXiv:2505.16373},
  year={2025}
}

@inproceedings{ddfm_2023_iccv,
  title={DDFM: Denoising diffusion model for multi-modality image fusion},
  author={Zhao, Zixiang and Bai, Haowen and Zhu, Yuanzhi and Zhang, Jiangshe and Xu, Shuang and Zhang, Yulun and Zhang, Kai and Meng, Deyu and Timofte, Radu and Van Gool, Luc},
  booktitle={Proceedings of the IEEE/CVF international conference on computer vision},
  pages={8082--8093},
  year={2023}
}

@ARTICLE{Mask_Difuser_2026_tpami,
  author={Tang, Linfeng and Li, Chunyu and Ma, Jiayi},
  journal={IEEE Transactions on Pattern Analysis and Machine Intelligence}, 
  title={Mask-DiFuser: A Masked Diffusion Model for Unified Unsupervised Image Fusion}, 
  year={2026},
  volume={48},
  number={1},
  pages={591-608},
  doi={10.1109/TPAMI.2025.3609323}}

@article{piafusion_2022_inf,
  title={PIAFusion: A progressive infrared and visible image fusion network based on illumination aware},
  author={Tang, Linfeng and Yuan, Jiteng and Zhang, Hao and Jiang, Xingyu and Ma, Jiayi},
  journal={Information Fusion},
  volume={83},
  pages={79--92},
  year={2022},
  publisher={Elsevier}
}

@inproceedings{mfnet_2017_iros,
  title={MFNet: Towards real-time semantic segmentation for autonomous vehicles with multi-spectral scenes},
  author={Ha, Qishen and Watanabe, Kohei and Karasawa, Takumi and Ushiku, Yoshitaka and Harada, Tatsuya},
  booktitle={2017 IEEE/RSJ International Conference on Intelligent Robots and Systems (IROS)},
  pages={5108--5115},
  year={2017},
  organization={IEEE}
}

@article{cvc14_sensors_2016,
  title={Pedestrian detection at day/night time with visible and FIR cameras: A comparison},
  author={Gonz{\'a}lez, Alejandro and Fang, Zhijie and Socarras, Yainuvis and Serrat, Joan and V{\'a}zquez, David and Xu, Jiaolong and L{\'o}pez, Antonio M},
  journal={Sensors},
  volume={16},
  number={6},
  pages={820},
  year={2016},
  publisher={MDPI}
}

@article{tno_2017,
  title={The TNO multiband image data collection},
  author={Toet, Alexander},
  journal={Data in brief},
  volume={15},
  pages={249},
  year={2017}
}

@inproceedings{KAIST_2015_cvpr,
  title={Multispectral pedestrian detection: Benchmark dataset and baseline},
  author={Hwang, Soonmin and Park, Jaesik and Kim, Namil and Choi, Yukyung and So Kweon, In},
  booktitle={Proceedings of the IEEE conference on computer vision and pattern recognition},
  pages={1037--1045},
  year={2015}
}

@inproceedings{freiburg_2020_iros,
  title={Heatnet: Bridging the day-night domain gap in semantic segmentation with thermal images},
  author={Vertens, Johan and Z{\"u}rn, Jannik and Burgard, Wolfram},
  booktitle={2020 IEEE/RSJ International Conference on Intelligent Robots and Systems (IROS)},
  pages={8461--8468},
  year={2020},
  organization={IEEE}
}

@inproceedings{LLVIP_2021_cvpr,
  title={LLVIP: A visible-infrared paired dataset for low-light vision},
  author={Jia, Xinyu and Zhu, Chuang and Li, Minzhen and Tang, Wenqi and Zhou, Wenli},
  booktitle={Proceedings of the IEEE/CVF international conference on computer vision},
  pages={3496--3504},
  year={2021}
}

@INPROCEEDINGS{havard_2011_cvpr,
  author={Chakrabarti, Ayan and Zickler, Todd},
  booktitle={CVPR 2011}, 
  title={Statistics of real-world hyperspectral images}, 
  year={2011},
  volume={},
  number={},
  pages={193-200},
  doi={10.1109/CVPR.2011.5995660}}

@inproceedings{icvl_2016_eccv,
  title={Sparse recovery of hyperspectral signal from natural RGB images},
  author={Arad, Boaz and Ben-Shahar, Ohad},
  booktitle={European conference on computer vision},
  pages={19--34},
  year={2016},
  organization={Springer}
}

@article{whu_hi_2020_arxiv,
  title={WHU-Hi: UAV-borne hyperspectral with high spatial resolution (H2) benchmark datasets for hyperspectral image classification},
  author={Hu, Xin and Zhong, Yanfei and Luo, Chang and Wang, Xinyu},
  journal={arXiv preprint arXiv:2012.13920},
  year={2020}
}

@inproceedings{arad_ntire_2022_cvpr,
  title={Ntire 2022 spectral recovery challenge and data set},
  author={Arad, Boaz and Timofte, Radu and Yahel, Rony and Morag, Nimrod and Bernat, Amir and Cai, Yuanhao and Lin, Jing and Lin, Zudi and Wang, Haoqian and Zhang, Yulun and others},
  booktitle={Proceedings of the IEEE/CVF Conference on Computer Vision and Pattern Recognition},
  pages={863--881},
  year={2022}
}

@ARTICLE{huston_2013_jstars,
  author={Debes, Christian and Merentitis, Andreas and Heremans, Roel and Hahn, Jürgen and Frangiadakis, Nikolaos and van Kasteren, Tim and Liao, Wenzhi and Bellens, Rik and Pižurica, Aleksandra and Gautama, Sidharta and Philips, Wilfried and Prasad, Saurabh and Du, Qian and Pacifici, Fabio},
  journal={IEEE Journal of Selected Topics in Applied Earth Observations and Remote Sensing}, 
  title={Hyperspectral and LiDAR Data Fusion: Outcome of the 2013 GRSS Data Fusion Contest}, 
  year={2014},
  volume={7},
  number={6},
  pages={2405-2418},
  doi={10.1109/JSTARS.2014.2305441}}

@misc{dai2026hadarbasedthermalinfraredhyperspectral,
  title         = {HADAR-Based Thermal Infrared Hyperspectral Image Restoration},
  author        = {Cheng Dai and Jiale Lin and Bingxuan Song and Yifei Chen and Jiashuo Chen and Xin Yuan and Fanglin Bao},
  year          = {2026},
  eprint        = {2605.13664},
  archivePrefix = {arXiv},
  url           = {https://arxiv.org/abs/2605.13664}
}

@article{2021_prl_xufeihu_non_line_of_sight_imaging,
  title={Non-line-of-sight imaging with picosecond temporal resolution},
  author={Wang, Bin and Zheng, Ming-Yang and Han, Jin-Jian and Huang, Xin and Xie, Xiu-Ping and Xu, Feihu and Zhang, Qiang and Pan, Jian-Wei},
  journal={Physical Review Letters},
  volume={127},
  number={5},
  pages={053602},
  year={2021},
  publisher={APS}
}

@article{ye2024real_yuanxin_xufeihu_ncs_2024,
  title={Real-time non-line-of-sight computational imaging using spectrum filtering and motion compensation},
  author={Ye, Jun-Tian and Sun, Yi and Li, Wenwen and Zeng, Jian-Wei and Hong, Yu and Li, Zheng-Ping and Huang, Xin and Xue, Xianghui and Yuan, Xin and Xu, Feihu and others},
  journal={Nature Computational Science},
  volume={4},
  number={12},
  pages={920--927},
  year={2024},
  publisher={Nature Publishing Group US New York}
}

@article{nature_machine_intelligence_2022_autonomous_driving,
  title={Deep learning-based robust positioning for all-weather autonomous driving},
  author={Almalioglu, Yasin and Turan, Mehmet and Trigoni, Niki and Markham, Andrew},
  journal={Nature machine intelligence},
  volume={4},
  number={9},
  pages={749--760},
  year={2022},
  publisher={Nature Publishing Group UK London}
}

@article{libradtran_software_emde2016libradtran,
  title     = {The libRadtran software package for radiative transfer calculations (version 2.0.1)},
  author    = {Emde, Claudia and Buras-Schnell, Robert and Kylling, Arve and Mayer, Bernhard and Gasteiger, Josef and Hamann, Ulrich and Kylling, Jonas and Richter, Bettina and Pause, Christian and Dowling, Timothy and others},
  journal   = {Geoscientific Model Development},
  volume    = {9},
  number    = {5},
  pages     = {1647--1672},
  year      = {2016},
  publisher = {Copernicus Publications G{\"o}ttingen, Germany}
}

@article{ALS_2025,
  title   = {Baseline correction with asymmetric least squares smoothing},
  author  = {Eilers, Paul H. C. and Boelens, Hans F. M.},
  journal = {Leiden University Medical Centre Report},
  volume  = {1},
  number  = {1},
  pages   = {5},
  year    = {2005}
}

@article{HySime_TGRS_2008,
  title     = {Hyperspectral subspace identification},
  author    = {Bioucas-Dias, Jos{\'e} M. and Nascimento, Jos{\'e} M. P.},
  journal   = {IEEE Transactions on Geoscience and Remote Sensing},
  volume    = {46},
  number    = {8},
  pages     = {2435--2445},
  year      = {2008},
  publisher = {IEEE}
}

@article{he_dehazing_2010_tpami,
  author  = {He, Kaiming and Sun, Jian and Tang, Xiaoou},
  title   = {Single Image Haze Removal Using Dark Channel Prior},
  journal = {IEEE Transactions on Pattern Analysis and Machine Intelligence},
  volume  = {33},
  number  = {12},
  pages   = {2341--2353},
  year    = {2011},
  doi     = {10.1109/TPAMI.2010.168}
}

@ARTICLE{vivnet_all_in_one_image_restoration_2026_tpami,
author={Cui, Yuning and Ren, Wenqi and Shi, Boxin and Knoll, Alois},
journal={ IEEE Transactions on Pattern Analysis \& Machine Intelligence },
title={{ Visual-in-Visual: A Unified and Efficient Baseline for Image Restoration }},
year={5555},
volume={},
number={01},
ISSN={1939-3539},
pages={1-18},
doi={10.1109/TPAMI.2026.3669720},
url = {https://doi.ieeecomputersociety.org/10.1109/TPAMI.2026.3669720},
publisher={IEEE Computer Society},
address={Los Alamitos, CA, USA},
month=mar}

@ARTICLE{wasserstein_all_in_one_image_restoration_2026_tpami,
author={Tang, Xiaole and He, Xiaoyi and Xu, Jiayi and Gu, Xiang and Sun, Jian},
journal={ IEEE Transactions on Pattern Analysis \& Machine Intelligence },
title={{ Learning Continuous Wasserstein Barycenter Space for Generalized All-in-One Image Restoration }},
year={5555},
volume={},
number={01},
ISSN={1939-3539},
pages={1-16},
doi={10.1109/TPAMI.2026.3669121},
url = {https://doi.ieeecomputersociety.org/10.1109/TPAMI.2026.3669121},
publisher={IEEE Computer Society},
address={Los Alamitos, CA, USA},
month=feb}

@article{zhuang2018fast,
  author  = {Zhuang, Lina and Bioucas-Dias, Jos{\'e} M.},
  title   = {Fast Hyperspectral Image Denoising and Inpainting Based on Low-Rank and Sparse Representations},
  journal = {IEEE Journal of Selected Topics in Applied Earth Observations and Remote Sensing},
  volume  = {11},
  number  = {3},
  pages   = {730--742},
  year    = {2018},
  doi     = {10.1109/JSTARS.2018.2796570}
}

@article{wangdu_isprs_2026_tes,
  title={Toward noise-resilient retrieval of land surface temperature and emissivity using airborne thermal infrared hyperspectral imagery},
  author={Wang, Du and Cao, Li-Qin and Du, Yu-Hao and Xiong, Hai-Yang and Ye, Fa-Wang and Zhong, Yan-Fei},
  journal={ISPRS Journal of Photogrammetry and Remote Sensing},
  volume={231},
  pages={532--551},
  year={2026},
  publisher={Elsevier}
}

@inproceedings{YellinConcurrentBandSelection2024,
  author    = {Yellin, Florence and McCloskey, Scott and Hill, Cole and Smith, Eric and Clipp, Brian},
  title     = {Concurrent Band Selection and Traversability Estimation From Long-Wave Hyperspectral Imagery in Off-Road Settings},
  booktitle = {Proceedings of the IEEE/CVF Winter Conference on Applications of Computer Vision},
  pages     = {7483--7492},
  year      = {2024}
}

@article{DorkenGallastegiAbsorptionBasedPassiveRange2025,
  author  = {Dorken Gallastegi, U. and Rueda-Chac{\'o}n, H. and Stevens, M. J. and Goyal, V. K.},
  title   = {Absorption-Based, Passive Range Imaging From Hyperspectral Thermal Measurements},
  journal = {IEEE Transactions on Pattern Analysis and Machine Intelligence},
  volume  = {47},
  number  = {5},
  pages   = {4044--4060},
  year    = {2025}
}

@article{BaoHeatassistedDetectionRanging2023,
  author  = {Bao, Fanglin and Wang, Xueji and Sureshbabu, Shree Hari and Sreekumar, Gautam and Yang, Liping and Aggarwal, Vaneet and Boddeti, Vishnu N. and Jacob, Zubin},
  title   = {Heat-Assisted Detection and Ranging},
  journal = {Nature},
  volume  = {619},
  number  = {7971},
  pages   = {743--748},
  year    = {2023},
  doi     = {10.1038/s41586-023-06174-6}
}

@misc{xu2026universalcomputationalthermalimaging,
  author       = {Xu, Hongyi and Wang, D. and Zhao, C. and Chen, J. and Lin, J. and Cao, L. and Zhong, Y. and She, Y. and Bao, Fanglin},
  title        = {Universal Computational Thermal Imaging Overcoming the Ghosting Effect},
  year         = {2026},
  eprint       = {2604.01542},
  archivePrefix= {arXiv},
  url          = {https://arxiv.org/abs/2604.01542}
}

@article{BaoWhyThermalImages2024,
  author  = {Bao, Fanglin and Jape, S. and Schramka, A. and Wang, J. and McGraw, T. E. and Jacob, Zubin},
  title   = {Why Thermal Images Are Blurry},
  journal = {Optics Express},
  volume  = {32},
  number  = {3},
  pages   = {3852--3865},
  year    = {2024}
}

@article{MousaPhysicsIntegratedInference2026,
  author  = {Mousa, Mohamed A. and Bauer, Leif and Yang, Ziyi and Singh, Utkarsh and Deka, Angshuman and Jacob, Zubin},
  title   = {Physics-Integrated Inference for Signal Recovery in Non-Gaussian Regimes},
  journal = {Applied Physics Letters},
  volume  = {128},
  number  = {17},
  pages   = {171101},
  year    = {2026},
  doi     = {10.1063/5.0324166},
  url     = {https://doi.org/10.1063/5.0324166}
}

@article{CaoLWIRHyperspectralImageClassification2022,
  title={LWIR hyperspectral image classification based on a temperature-emissivity residual network and conditional random field model},
  author={Cao, Liqin and He, Jiani and Gao, Lyuzhou and Zhong, Yanfei and Hu, Xin and Li, Zhijiang},
  journal={International Journal of Remote Sensing},
  volume={43},
  number={10},
  pages={3744--3768},
  year={2022},
  publisher={Taylor \& Francis}
}

@inproceedings{BijelicSeeingThroughFog2020,
  author    = {Bijelic, Mario and Gruber, Tobias and Mannan, Fahim and Kraus, Florian and Ritter, Werner and Dietmayer, Klaus and Heide, Felix},
  title     = {Seeing Through Fog Without Seeing Fog: Deep Multimodal Sensor Fusion in Unseen Adverse Weather},
  booktitle = {Proceedings of the IEEE/CVF Conference on Computer Vision and Pattern Recognition},
  pages     = {11682--11692},
  year      = {2020}
}

@inproceedings{ParkRethinkingLiDARWeather2024,
  author    = {Park, Junsung and Kim, Kyungmin and Shim, Hyunjung},
  title     = {Rethinking Data Augmentation for Robust LiDAR Semantic Segmentation in Adverse Weather},
  booktitle = {Computer Vision -- ECCV 2024},
  pages     = {320--336},
  year      = {2024},
  publisher = {Springer},
  doi       = {10.1007/978-3-031-72640-8_18}
}

@article{TsengBlockCoordinateDescent2001,
  author  = {Tseng, Paul},
  title   = {Convergence of a Block Coordinate Descent Method for Nondifferentiable Minimization},
  journal = {Journal of Optimization Theory and Applications},
  volume  = {109},
  number  = {3},
  pages   = {475--494},
  year    = {2001},
  doi     = {10.1023/A:1017501703105}
}

\end{document}